\documentclass[12pt]{article}
\usepackage[a4paper,top=2cm,bottom=2cm,left=3cm,right=3cm,marginparwidth=1.75cm]{geometry}
\usepackage{graphicx}
\usepackage{amsmath,amssymb}
\usepackage[giveninits=true,sorting=none,style=numeric,minbibnames=5,maxbibnames=5]{biblatex}
\usepackage{csquotes}
\usepackage{url}
\usepackage{hyperref}
\usepackage{dsfont}
\usepackage{graphicx}
\usepackage{caption}
\usepackage{subcaption}
\usepackage{tikz}
\newif\ifreview
\reviewfalse

\ifreview
	\usepackage{lineno}

	\linenumbers
\fi

\title{\vspace{-2.5cm}{\bf Strategies for training point distributions in physics-informed neural networks}}
\author{Santosh Humagain, Toni Schneidereit \\ 
\\[1ex]
\small Chair for Applied Mathematics\\ 
\small Brandenburg University of Technology Cottbus-Senftenberg \\
\small Platz der Deutschen Einheit 1, 03046 Cottbus, Germany \\
\small \{Santosh.Humagain, Toni.Schneidereit\}@b-tu.de
}
\date{} 

\begin{document}

\maketitle

\begin{abstract}
Physics-informed neural networks approach the approximation of differential equations by directly incorporating their structure and given conditions in a loss function. This enables conditions like, e.g., invariants to be easily added during the modelling phase. In addition, the approach can be considered as mesh free and can be utilised to compute solutions on arbitrary grids after the training phase.  Therefore, physics-informed neural networks are emerging as a promising alternative to solving differential equations with methods from numerical mathematics. However, their performance highly depends on a large variety of factors. 
In this paper, we systematically investigate and evaluate a core component of the approach, namely the training point distribution. We test two ordinary and two partial differential equations with five strategies for training data generation and shallow network architectures, with one and two hidden layers. In addition to common distributions, we introduce sine-based training points, which are motivated by the construction of Chebyshev nodes.
The results are challenged by using certain parameter combinations like, e.g., random and fixed-seed weight initialisation for reproducibility. The results show the impact of the training point distributions on the solution accuracy and we find evidence that they are connected to the characteristics of the differential equation.
\end{abstract}
\begin{center}
\vspace{0.875cm}
\end{center}

\section{Introduction}

Differential Equations (DEs) are mathematical tools for modelling dynamic processes across various disciplines.  They enable us to describe, understand, and investigate physical processes, where quantities change over time or space \cite{boyce2017elementary}. For instance, exponential decay of radioactive substances, which can be described by a first-order ordinary differential equation (ODE), helps us to predict remaining substances at any given time and allows us to identify the age of ancient artefacts through carbon dating \cite{young2012sears}. 
Likewise, the Poisson equation, an inhomogeneous partial differential equation (PDE), is used to describe, e.g.,  an electrostatic field with a source term \cite{arfken2012mathematical}.

Despite the effectiveness of traditional analytical and numerical methods in solving differential equations, these methods often encounter significant challenges when solving complex, high-dimensional, or nonlinear problems \cite{boyce2017elementary}. Analytical methods, while yielding exact solutions and having the ability to give deep insight into variables \cite{boyce2017elementary}, become unwieldy or inapplicable for most engineering and scientific problems of practical interest \cite{strauss2007partial}. Numerical methods, such as finite difference and finite element methods, are of great use in approximating the solutions of differential equations. However, Chapra and Canale (2014) highlight that these methods are limited by truncation and round-off errors, which can affect stability and convergence. Moreover, when applied to the stiff equations or problems with complex boundary conditions, computational cost and error propagation can be significant concerns \cite{canale2014numerical}.

As an alternative approach, Lagaris et al. proposed a neural‑network collocation scheme in which the trial solution is written as the sum of a term that exactly satisfies the initial or boundary conditions and a feed‑forward neural network whose adjustable parameters are optimised so the overall expression fulfils the governing differential equation, yielding a differentiable closed‑form approximation \cite{lagaris1998artificial}. In recent years, the development of machine learning techniques, particularly physics-informed neural networks (PINNs), has opened new avenues to solve differential equations in mesh-free manners. PINNS integrate the governing physical principles directly into the training process of deep neural networks, thereby eliminating many of the limitations associated with traditional discretisation methods \cite{raissi2019physics}. Despite the widespread recognition of the potential of PINNs, there remain gaps in understanding how factors such as the distribution of grid (training) points in the domain, grid sizes, network architecture, and weight initialisation jointly influence their performance.


\paragraph{\textit{Literature review}}

In 1998, Lagaris et al. \cite{lagaris1998artificial} presented a novel method for solving ordinary and partial differential equations using artificial neural networks (ANNs). The approach constructs a trial solution as the sum of two parts: one that satisfies exactly the boundary or initial conditions without having any adjustable parameters, and another involving a feed-forward neural network whose weights are optimised to minimise the residual of the differential equation. This formulation transforms a constrained optimisation problem into an unconstrained one, simplifying the solution process considerably and giving a closed-form, differentiable solution with very good generalisation across the domain. Later in 2000, Lagaris et al. \cite{lagaris2000neural} used this ANN framework to solve boundary value problems with irregular boundaries. They combined ANNs and radial basis function networks to enforce complex boundary conditions. These works highlight the applicability of artificial neural network methods to solve a wide range of problems. 

The foundational ANN approach was extended by Raissi et al. \cite{raissi2019physics} in 2019 with embedding physical laws describing differential equations directly into the loss function formulation of a neural network to form physics-informed neural networks. These are a novel machine learning approach that directly integrates physical laws during the neural network training. Let us note that the foundations of this approach have already been mentioned by Lagaris et al. \cite{lagaris2000neural} in the year 2000. However, with the contribution of Raissi et al. \cite{raissi2019physics} in 2019, the approach gained more attention. 

Recent advancements have shown that PINNs are capable of solving both forward as well as inverse and parametric problems. Cuomo et al. \cite{cuomo2022scientific} state that PINNs have been successfully applied to challenging domains such as fluid dynamics and astrophysics. The mesh-free nature of PINN efficiently handles complex geometries and high-dimensional issues. They also highlight that PINNs can be sensitive to hyperparameter choices and the distribution of collocation points, which directly influences convergence and the minimisation of PDE residuals.

Baty \cite{baty2024hands} explains that PINNs provide a flexible, hands-on treatment of the solution of partial differential equations by directly including the boundary conditions, either as soft constraints via training data or as hard constraints via special trial functions, within the neural network architecture. This method eliminates the burden of discretisation of traditional numerical methods and also accommodates diverse boundary conditions and complex geometries. These advantages make PINNs a good choice for solving inverse and parametric problems with limited data to overcome the computational challenges of conventional methods.


Distributions of training points are important in PINN training, determining accuracy and convergence. Traditional approaches often utilise fixed sampling methods. These kinds of distributions are easy to implement, but often lead to poor accuracy as they may ignore the characteristics of solutions to PDEs \cite{anadaptiveliu2024}. Breuß et al. \cite{breuss2016numerical} showed that Fast Leja points match the interpolation accuracy similar to Chebyshev nodes but with lower computational cost. This research also discusses the ability of Chebyshev nodes over equidistant nodes to approximate the Runge function \cite{Epperson01041987}.

Guo et al. \cite{guo2022novel} presented an adaptive sampling technique for PINN. This technique weights PDE residuals in each time slice by a causality factor which enforces natural time ordering and boosts accuracy. Recent work by Fang et al. \cite{fang2023physics} presents a PINN framework which uses a mixed data sampling method, which combines Cartesian grid sampling with Latin hypercube sampling. This approach significantly improves the efficiency and accuracy in solving the forward and backwards modified diffusion equations by optimising the distribution of collocation points, especially under periodic boundary conditions . Going one step further, recent advances have proposed adaptive sampling methods that improve the distribution of collocation points based on error indicators. For example, Liu et al. \cite{anadaptiveliu2024} introduced an adaptive sampling method that leverages an expected improvement function in combination with residual gradient information to allocate more points dynamically to regions where PINN errors are significant, enhancing both accuracy and computational cost. Also, Daw et al. \cite{rethinking_daw2022} provided a critical examination of the fixed and purely residual-based sampling methods , which further justifies the need for investigating existing grid distribution techniques and/or finding new grid distribution strategies.

Recent advances in the optimisation of PINNs have emphasised the role of network architecture in determining accuracy as well as computational cost. For instance, Kaplarević-Mališić et al. \cite{kaplarevic2023identifying} demonstrated that even a small change in network architecture, such as variations in the number of layers and neurons per layer, can significantly affect the model performance. While his research was focused on an automated architecture design approach using an evolution strategy, he has also suggested a promising direction is to search for architecture design with fewer layers with a larger number of neurons. 

Existing research also shows that weight initialisation plays a crucial role in the convergence and stability of neural network solvers for differential equations \cite{schneidereit2022computational}. Batuwatta-Gamage et al. \cite{batuwatta2023weight} show that PINNs are prone to make inconsistent predictions and suffer from loss convergence issues when weights are randomly initialised, but this can be mitigated by nearly optimised, transfer-learning-based initialisation. On the other hand, deterministic weight initialisation, for example, setting all the weights to zero, doesn’t break symmetry and is likely to lead to a trivial or non-convergent solution. In addition to this, Saratchandran et al. \cite {saratchandran2024activation} offer a solid theoretical framework demonstrating that proper random initialisation schemes, such as Glorot Uniform, Xavier, or Kaiming, can significantly reduce the over-parameterisation required for gradient descent to converge effectively. 
 
\paragraph{\textit{Problem statement and contribution}}

There have been studies on sampling strategies for PINNs to improve their accuracy and performance. However, there has not  been a comparative study on the accuracy and performance of PINNs on commonly used random and deterministic spacing training points yet. This paper seeks to address both common and uncommon training point distribution strategies. In our study, those points are distributed equidistantly, in a random and random sorted order, based on Chebyshev nodes and our proposed sine-based strategy.
In addition to this, weight initialisation and network architecture are found to have a significant influence on the accuracy and performance of PINNs. Only a few studies have explored their combined effect yet. This paper also tries to compare the performance of these training point distributions with two different network architectures and different weight initialisation schemes, random and deterministic. For testing purposes, we apply the PINN framework to two ordinary and two partial differential equations. 
 
The paper is structured as follows. After an introduction to the PINN framework with the feed-forward neural network, the loss function and the optimisation, we introduce the differential equations and the training point distribution strategies. This is followed by the utilised performance evaluation metrics and finally we present the results, discussions and conclusions.

\section{The PINN framework}

The physics-informed neural network framework in our paper consists three parts: \textit{(i)} the utilised neural network, \textit{(ii)} the loss function based on the differential equation with given conditions, and \textit{(iii)} the neural network weight optimisation. 

\subsection{Feed-forward neural network}
\label{Sec:ANN}
\begin{figure}[!ht]
\centering
\includegraphics[scale=1]{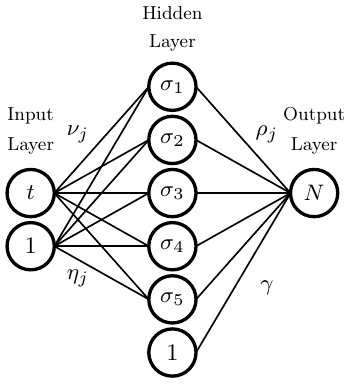}
\caption{Visualisation of an example feed-forward neural network with five hidden neurons and a bias neuron in both the input and hidden layer. ($t$) input neuron, ($\sigma_j$) hidden neurons, ($1$) bias neuron, ($N$) neural network output, ($\nu_j,\eta_j,\rho_j,\gamma$) adjustable weights.}
\label{fig:FNN}
\end{figure}

The ANN in the form of a feed-forward neural network (FNN) is an essential building block in PINNs, where information flows in a single direction: from the input layer through one or more hidden layers to the output layer. To describe how a FNN can make predictions based on a given input, the diagram shown in Figure \ref{fig:FNN} is taken into consideration \cite {schneidereit2022solution}.

\textbf{Input Layer:} 
The input layer receives the discretised data (e.g., spatial or temporal training points) from the problem domain. For the description of the FNN, here the temporal coordinate $t$ in the domain $D$ is taken. A bias neuron with a fixed value of 1 is often added as a flexible offset in subsequent layers. The number of input neurons is determined by the number of independent variables. 

\textbf{Hidden Layer:} 
The hidden layer performs a weighted summation of inputs, followed by a non-linear activation function. Considering the hidden layer consists of $H$ hidden neurons $\sigma_j,~j=1,\ldots,H$, their input is given by $z_{j} = \nu_{j}t + \eta_{j}$. Here, $\nu_{j}$ are the weights for the discrete input $t$, and $\eta_{j}$ are the weights for the input bias neuron. This input is then passed through a non-linear activation function $\sigma_j=\sigma\left(z_{j}\right) =\sigma\left(\nu_{j}t + \eta_{j}\right)$. The activation function used in this work is the hyperbolic tangent $tanh(z_{j})$ function. This function maps any real argument to the bounded, zero-centered interval of $[-1,1]$. The continuously differentiable nature helps to maintain stable gradients and effectively capture the continuous dynamics present in differential equations \cite{dubey2022activation,baty2024hands}.

\textbf{Output Layer:} 
The output layer aggregates the hidden layer outputs and combines them with another set of weights and a bias term to produce the final network output, which is given by
\begin{equation}
    N_{\theta}(t) = \sum_{j=1}^{H}{\rho_{j}\sigma\left( z_{j} \right) + \gamma}
    \label{eq:FNNoutput}
\end{equation}
In Eq.\ \eqref{eq:FNNoutput}, $\sigma\left( z_{j} \right)$ are outputs of hidden layer neurons, connected with weights $\rho_{j}$ and a bias neuron with the weight $\gamma$. The activation function in the output layer is linear and all the FNN weights $\theta=(\nu_j, \eta_j, \rho_j, \gamma)$ are now subject to optimisation in order to find accurate predictions. 

\subsection{PINN loss function and optimisation}
\label{Sec:PINNloss}

PINNs \cite{raissi2019physics} combine deep learning with the physical laws that are described by differential equations. While traditional machine learning models depend entirely on data, PINNs embed physics-based constraints into the learning process. This makes PINNs suitable for solving both forward and inverse problems in dynamic systems. Following Raissi et al. \cite{raissi2019physics}, the target PDE is written in the generic form as,
\begin{equation}
u_t(t,\vec{x}) + \mathcal{N}[u(t,\vec{x})] = 0,~~\vec{x}\in\Omega,~t\in[0,T] 
\label{eq:targetPDE}
\end{equation}
with given initial condition $u(0,\vec{x})=u_{\text{initial}}(\vec{x})$ and boundary condition $u(t,\vec{x})=u_{\text{bound}}(t,\vec{x}),~\vec{x}\in\partial\Omega$. In Eq.\ \eqref{eq:targetPDE}, $\vec{x}$ represents the spatial coordinates in the domain $\Omega\subset\mathbb{R}^d$. Depending on the problem, this can include several directions, e.g., $x,y,z$. The temporal coordinate is given by $t$. While $u(t,\vec{x})$ denotes the exact solution, the term $\mathcal{N}[u(t,\vec{x})]$ is a differential operator that represents the spatial components of the PDE. 

PINNs aim to approximate the exact solution $u(t,\vec{x})$ by using a neural network output $u_\theta(t,\vec{x})$, where $\theta$ represents the neural network weights and biases. In Sec.\ \ref{Sec:ANN}, we introduced $N_{\theta}(t)$ as the output of a purely temporal FNN that takes only $t$ as input. In the PINN formulation, we generalise the same network architecture to the number of inputs defined by $\vec{x}$ and $t$, writing $u_{\theta}(t,\vec{x})$.
Thus, $N_{\theta}(t)$ becomes $N_{\theta}(t,\vec{x})$ and resembles $u_{\theta}(t,x)$ as the same underlying neural map. The loss function in the PINN framework ensures that the solution satisfies both the PDE and the given intial/boundary conditions \cite{lagaris1998artificial, raissi2019physics, schneidereit2022computational}. The combined PINN loss is a composition of the residual loss $\mathcal{L}_{\mathcal{R}}$, the initial condition loss $\mathcal{L}_{\mathcal{IC}}$, and the boundary condition loss $\mathcal{L}_{\mathcal{BC}}$. Given that $\vec{x}=x$ is one-dimensional in the following example, the total loss is defined as \cite{raissi2019physics,Wang2023AnEG}:
\begin{equation}
    \mathcal{L}(\theta) = \mathcal{L}_{\mathcal{R}}(\theta) + \mathcal{L}_{\mathcal{IC}}(\theta) + \mathcal{L}_{\mathcal{BC}}(\theta)
\end{equation}
with mean squared error (MSE) terms 
\begin{equation}
    \left\{
    \begin{array}{l}
    \displaystyle \mathcal{L}_{\mathcal{R}}(\theta) = \frac{1}{N_t^{\mathcal{R}}N_x^{\mathcal{R}}}\sum_{n = 1}^{N_t^{\mathcal{R}}}\sum_{i = 1}^{N_x^{\mathcal{R}}}\left| u_{\theta,t}(t_n^{\mathcal{R}},x_i^{\mathcal{R}}) + \mathcal{N}\left[ u_{\theta}(t_n^{\mathcal{R}},x_i^{\mathcal{R}})\right] \right|^{2} \\[0.4cm]
    \displaystyle \mathcal{L}_{\mathcal{IC}}(\theta) = \frac{1}{N_x^{\mathcal{IC}}} \sum_{i=1}^{N_x^{\mathcal{IC}}} \left| u_\theta(0, x_i^{\mathcal{IC}}) - u_{\text{initial}}(x_i^{\mathcal{IC}}) \right|^2 \\[0.4cm]
    \displaystyle \mathcal{L}_{\mathcal{BC}}(\theta) = \frac{1}{N_t^{\mathcal{BC}}N_x^{\mathcal{BC}}} \sum_{n=1}^{N_t^{\mathcal{BC}}}\sum_{i=1}^{N_x^{\mathcal{BC}}} \left| u_\theta(t_n^{\mathcal{BC}}, x_i^{\mathcal{BC}}) - u_{\text{bound}}(t_n^{\mathcal{BC}},x_i^{\mathcal{BC}}) \right|^2 
    \end{array}\right.
    \label{eq:total_loss}
\end{equation}

\begin{table}[!h] 
\centering
\caption{Overview of the training sets for the residual loss and the initial/boundary condition loss in Eq.\ \eqref{eq:total_loss}. \label{tab5}}
\begin{tabular}{l|l|l}
\textbf{loss part}	& \textbf{training set}	& \textbf{range}\\ 
\hline
Differential equation &$t_n^{\mathcal{R}},x_i^{\mathcal{R}}$ &  $n=1,\ldots,N_t^{\mathcal{R}}$   \\[0.1cm]
& & $i=1,\ldots,N_x^{\mathcal{R}}$ \\[0.25cm]
Initial condition	& $x_i^{\mathcal{IC}}$ & $i=1,\ldots,N_x^{\mathcal{IC}}$\\[0.25cm]
Boundary condition & $t_n^{\mathcal{BC}},x_i^{\mathcal{BC}}$ & $n=1,\ldots,N_t^{\mathcal{BC}}$ \\[0.1cm]
& & $i=1,\ldots,N_x^{\mathcal{BC}}$ \\
\end{tabular}
\end{table}

In Eq.\ \eqref{eq:total_loss}, the model computes $u_{\theta,t}(t,x)$, which is the time derivative of the ANN output $u_{\theta}(t,x)$ with respect to $t$, by using automatic differentiation \cite{AutoDiff, JMLR:v18:17-468}.

The expression $\mathcal{L}_{\mathcal{R}}$ penalises the MSE of the residual of the differential equation, ensuring that the FNN prediction satisfies the PDE structure throughout the domain. Minimising this term enforces consistency with the governing equation at every training point. The same holds for the initial condition loss $\mathcal{L}_{\mathcal{IC}}$ and boundary condition loss $\mathcal{L}_{\mathcal{BC}}$.

In order to effectively minimise the total loss and the neural network to learn a valid approximation of the PDE, we use Adam (adaptive moment estimation) \cite{kingma2014adam} optimisation. This is a first-order stochastic optimisation algorithm that automatically adjusts the ANN weights and biases by tracking the mean and the variance of past gradients. At iteration $k$, we compute the gradient $g_k = \nabla_\theta \mathcal{L}(\theta_{k-1})$ and update two moving averages:
\begin{equation}
m_k = \beta_1\,m_{k-1} + (1 - \beta_1)\,g_k,~~
v_k = \beta_2\,v_{k-1} + (1 - \beta_2)\,g_k^2    
\end{equation}
where $\beta_1,\beta_2$ (both between 0 and 1) control how much impact previous gradients have in order to not get stuck in a shallow minimum. In addition, those parameters control how quickly the algorithm forgets old gradient information. Since $m_k$ and $v_k$ are initialised with zeros (and no estimation of previous gradients), they are biased towards zero during the first few steps. To compensate for this, Adam computes bias-corrected estimates
\begin{equation}
\widehat m_k = \frac{m_k}{1 - \beta_1^k},~~
\widehat v_k = \frac{v_k}{1 - \beta_2^k}    
\end{equation}
The parameters are then updated via
\begin{equation}
\theta_k = \theta_{k-1} - \alpha\,\frac{\widehat m_k}{\sqrt{\widehat v_k} + \varepsilon}    
\end{equation}
where $\alpha$ is the initial learning rate and $\varepsilon$ is a small constant added for numerical stability. The hyperparameters $\alpha = 10^{-3}, \beta_1 = 0.9, \beta_2 = 0.999,\varepsilon = 10^{-8}$ are found to be robust by the authors of Adam \cite{kingma2014adam}.

\section{Methodology}

\label{sec: methodology}
In this section, we introduce the two ODEs and two PDEs utilised for our experiments, as well as the training point distribution strategies, the evaluation metrics and experimental design.

\subsection{Benchmark equations}

\begin{itemize}
\item \textbf{Radioactive decay:}
Radioactive decay is modelled by a first-order differential equation \cite{young2012sears}:

\begin{equation}
\frac{dx(t)}{dt} = - \lambda x(t),~~ x(0) = x_{0}
\end{equation}
where $x(t)$ represents the number of radioactive nuclei at time $t$, $x_{0}$ is the initial quantity and $\lambda$ is the decay rate, which is positive. The analytical solution is given by
\begin{equation}
    x(t) = x_{0}e^{- \lambda t}
\end{equation}
To enforce that the residual is minimised over the domain, the residual loss and the initial condition loss, which ensures that the solution satisfies $x(0) = x_0$, are given by:
\begin{equation}
    \mathcal{L} = \frac{1}{N_t^{\mathcal{R}}}\sum_{n=1}^{N_t^{\mathcal{R}}}\left| x_{\theta,t}(t_n^{\mathcal{R}}) + \lambda\, x_{\theta}(t_n^{\mathcal{R}}) \right|^2+\left| x_{\theta}(0) - x_0 \right|^2
\end{equation}

\item \textbf{Simple harmonic oscillator:} The simple harmonic oscillator is a second-order ordinary differential equation used to describe oscillatory behaviours in electrical and mechanical systems. The dynamics of a simple harmonic oscillator are given by \cite{young2012sears}:
\begin{equation}
\frac{d^2 x(t)}{dt^2} + \omega^2 x (t) = 0,~x(0)=x_0,~x_t(0)=v_0
\label{eq:SHO_initial}
\end{equation}
where $\omega$ is the angular velocity. This equation represents the motion of a system where the restoring force is proportional to the displacement, and it exhibits periodic behaviour. The analytical solution to this equation is:
\begin{equation}
   x(t)=x_0 \cos(\omega t)+\frac{v_0}{\omega} \sin(\omega t) 
\end{equation}
The residual loss and the initial condition loss are combined as:
\begin{equation}
    \mathcal{L} = \frac{1}{N_t^{\mathcal{R}}}\sum_{n=1}^{N_t^{\mathcal{R}}}\left| x_{\theta,tt}(t_n^{\mathcal{R}}) + \omega^2\, x_{\theta}(t_n^{\mathcal{R}}) \right|^2 + \left| x_{\theta}(0) - x_0 \right|^2 + \left| x_{\theta,t}(0) - v_0 \right|^2
\end{equation}

\item \textbf{Laplace equation:} The Laplace equation is a classical elliptic PDE used to model steady-state phenomena such as electrostatic potentials and incompressible fluid flow. The two-dimensional Laplace equation is defined as \cite{baty2024hands}:
\begin{equation}
\left\{
\begin{array}{l}
\displaystyle \frac{\partial^{2}u(x,y)}{{\partial x}^{2}} + \frac{\partial^{2}u(x,y)}{{\partial y}^{2}} = 0,~(x,y) \in \Omega \\[0.1cm]
u( - 1,y) = u(1,y) = 1 - y^{2},~\text{vertical boundary} \\[0.1cm]
u(x, - 1) = u(x,1) = x^{2} - 1,~\text{horizontal boundary}
\end{array}\right.
\label{eq:Laplace_PDE_boundary}
\end{equation}
These Dirichlet boundary conditions guarantee the uniqueness of the solution and serve as a strong constraint for the PINN during training. In our study, we consider a square domain $\Omega = [-1,1] \times [-1,1]$. The exact solution for the Laplace equation is then given by:
\begin{equation}
u(x,y) = x^{2} - y^{2}
\end{equation}
We discretise the interior of $\Omega$ by choosing $N_x^{\mathcal{R}}$ sample points in the $x$-direction and $N_y^{\mathcal{R}}$ sample points in the $y$-direction. These form a Cartesian grid of $N_x\times N_y$ training points $(x_i^{\mathcal{R}},y_j^{\mathcal{R}}),, i=1,\ldots,N_x^{\mathcal{R}}, j=1,\ldots,N_y^{\mathcal{R}}$. For the boundary loss, we use the values of the given conditions $u_{\text{bound}}(x_i^{\mathcal{BC}},y_j^{\mathcal{BC}})$ with discrete points $(x_i^{\mathcal{BC}},y_j^{\mathcal{BC}}), i=1,\ldots,N_x^{\mathcal{BC}}, j=1,\ldots,N_y^{\mathcal{BC}}$ on $\partial\Omega$. The combined residual and boundary loss is then given as
\begin{align}
    \mathcal{L} = &\frac{1}{N_{x}^{\mathcal{R}}N_{y}^{\mathcal{R}}}\sum_{i=1}^{N_{x}^{\mathcal{R}}}\sum_{j=1}^{N_{y}^{\mathcal{R}}}\left| u_{\theta,xx}(x_i^{\mathcal{R}},y_j^{\mathcal{R}}) + u_{\theta,yy}(x_i^{\mathcal{R}},y_j^{\mathcal{R}}) \right|^2 \\ \nonumber
    &+ \frac{1}{N_{x}^{\mathcal{BC}}N_{y}^{\mathcal{BC}}}\sum_{i=1}^{N_{x}^{\mathcal{BC}}}\sum_{j=1}^{N_{y}^{\mathcal{BC}}}\left| u_\theta(x_i^{\mathcal{BC}},y_j^{\mathcal{BC}}) - u_{\text{bound}}(x_i^{\mathcal{BC}},y_j^{\mathcal{BC}}) \right|^2
\end{align}

\item \textbf{Poisson equation:} The Poisson equation is a generalisation of the Laplace equation that includes a source term. In two dimensions, the Poisson equation is given by \cite{baty2024hands}:

\begin{equation}
\left\{
\begin{array}{l}
\displaystyle \frac{\partial^{2}u(x,y)}{{\partial x}^{2}} + \frac{\partial^{2}u(x,y)}{{\partial y}^{2}} = \left(x^2+y^2\right)e^{xy},~ (x,y) \in \Omega  \\[0.1cm]
u(0,y) = 1,~u(1,y) = e^{y},~\text{vertical boundary} \\[0.1cm]
 u(x,0) = 1, u(x,1) = e^{x},~\text{horizontal boundary}
\end{array}\right.
\label{eq:Laplace_PDE_boundary}
\end{equation}
with $\Omega = [0,1] \times [0,1]$. The exact solution is given by
\begin{equation}
u(x,y) = e^{xy}
\end{equation}
The discretisation follows the same procedure as described for the Laplace equation. The combined loss then reads
\begin{align}
\mathcal{L} = &\frac{1}{N_{x}^{\mathcal{R}}N_{y}^{\mathcal{R}}}\sum_{i=1}^{N_{x}^{\mathcal{R}}}\sum_{j=1}^{N_{y}^{\mathcal{R}}}\left| u_{\theta,xx}(x_i^{\mathcal{R}},y_j^{\mathcal{R}}) + u_{\theta,yy}(x_i^{\mathcal{R}},y_j^{\mathcal{R}}) - f(x_i^{\mathcal{R}},y_j^{\mathcal{R}}) \right|^2 \\ \nonumber
&+ \frac{1}{N_{x}^{\mathcal{BC}}N_{y}^{\mathcal{BC}}}\sum_{i=1}^{N_{x}^{\mathcal{BC}}}\sum_{j=1}^{N_{y}^{\mathcal{BC}}}\left| u_\theta(x_i^{\mathcal{BC}},y_j^{\mathcal{BC}}) - u_{\text{bound}}(x_i^{\mathcal{BC}},y_j^{\mathcal{BC}}) \right|^2
\end{align}

\end{itemize}

\subsection{Training point distribution strategies}
\label{training_point_generation}

In this section, we discuss the various strategies used to generate the training point distributions for the PINN training of our benchmark equations. For interior and boundary data, the following grid types are considered:

\begin{itemize}
    \item \textbf{Equidistant grid:} The training points are evenly distributed over the domain $[a,b]$.
    \item \textbf{Random grid:} The training points are independently sampled from a uniform distribution over the domain $[a,b]$. This produces the scattered distribution of points over the domain. The training points are used in their order of appearance.
    \item \textbf{Random sorted grid:} The training points are first generated randomly over the domain $[a,b]$ and then sorted in ascending order. Although the spacing is still non uniform, sorting introduces a different order in the training domain.
    \item \textbf{Chebyshev grid:} The training points are generated as per Chebyshev nodes over the domain $[a,b]$ and are given by \cite{han2020better, breuss2016numerical}:
    \begin{equation}
        t_i = \frac{a+b}{2} + \frac{b-a}{2} \cos\!\left(\frac{2i+1}{2n}\pi\right),~i = 0,\ldots, n-1
    \end{equation}
    This results in clustering of points near the boundaries of the domain $[a,b]$.

    \item \textbf{Sine-based grid:} The training point distribution that we propose along the paper is generated based on the arc length of a full sine wave with its amplitude half of the training domain. One can imagine this as a complete sine wave in a square box whose length is equal to the domain:
    \begin{equation}
    \left\{
    \begin{array}{l}
    \displaystyle l_i = \frac{i}{n-1} L_{arc}, \quad i = 0,\ldots,n-1, \\[0.3cm]
    \displaystyle L_{arc} = \int_{a}^{b} \sqrt{1 + \left( \frac{b - a}{2} \cdot \frac{2\pi}{b - a} \cos\left( \frac{2\pi (s - a)}{b - a} \right) \right)^2 } \, ds,
    \end{array}
    \right.
    \end{equation}
    This procedure yields a non-uniform distribution of training points, where the density is controlled by the curvature of the sine wave.

\end{itemize}
 
\begin{figure}[!ht]
\centering
\includegraphics[scale=0.5]{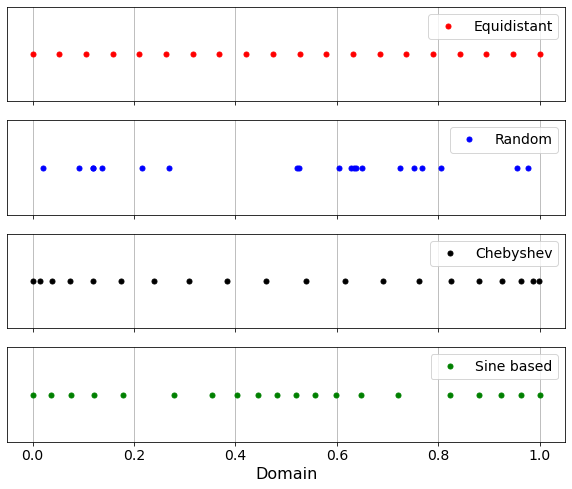}
\caption{Visualisation of the training point distribution strategies in a domain $[0,1]$ using 20 grid points.}
\label{fig:grid_distribution}
\end{figure}

The plot in Fig. \ref{fig:grid_distribution} compares the distribution of 20 training points on the domain $[0,1]$ with four different techniques: equidistant, random, Chebyshev, and sine-based. The equidistant grid (red) provides equally spaced points, while the random grid offers a scattered, unstructured distribution that will vary from run to run; its sorted version (random sorted) would have an ascending ordering, a distinction not shown in the figure. The Chebyshev grid (black) clusters points towards the boundaries, and our sine-based grid (green) spaces points according to equal arc-length along a sine wave, resulting in non-uniform spacing that follows the geometry of the curve. The comparison demonstrates how various sampling strategies can result in varied spatial and temporal distributions of training points.

\subsection{Evaluation metrics}
For the assessment of the PINN performance, we utilise standard evaluation metrics that compare the PINN predictions ($u_\theta$) with the exact solution ($u_\text{exact}$) of our benchmark equations. We also consider the following evaluation metrics:

\begin{itemize}
    \item \textbf{Mean absolute error (MAE):} MAE represents the average magnitude of the prediction errors. A smaller MAE value indicates that the predictions deviate less from the true values. The formula is given by:    
    \begin{equation}
        \text{MAE} = \frac{1}{N} \sum_{i=1}^{N} \left| u_{\theta, i} - u_{\text{exact}, i} \right|
    \end{equation}

    \item \textbf{Average ($\overline{\text{\textbf{MAE}}}$):} When the random weight initialisation is used, every individual (complete) optimisation may return a different MAE. In such cases, we compute the average of MAE over several runs to see the average accuracy. If $\{MAE_i\}_{i=1}^O$ are the MAE values over $O$ independent runs, then the average of those errors is defined as:
    \begin{equation}
        \overline{\text{MAE}} = \frac{1}{O} \sum_{i=1}^{O}\text{MAE}_i
    \end{equation}
    \item \textbf{Standard deviation (SD):}
    To quantify the variability of the MAE across $O$ independent (complete) optimisations, we compute its standard deviation. If $\{\text{MAE}_i\}_{i=1}^O$ are the individual error values and $\overline{\text{MAE}}$ is their mean, then the standard deviation is defined as
    \begin{equation}
        \text{SD} = \sqrt{\frac{1}{O}\sum_{i=1}^O\bigl(\text{MAE}_i - \overline{\text{MAE}}\bigr)^2}
    \end{equation}


\end{itemize}

\subsection{Experimental design}

The experiments are designed to give a robust reference point for the PINN approach by solving four benchmark differential equations \textit{(i)} radioactive decay, \textit{(ii)} simple harmonic oscillator, \textit{(iii)} Laplace equation, \textit{(iv)} Poisson equation. 

We utilise a fully connected FNN with a single hidden layer with 100 neurons as one configuration and two hidden layers with 50 neurons as a second configuration. The input and output layers both use a linear activation, while the hidden layer uses the hyperbolic tangent activation function. The FNN weights are trained with Adam optimisation and hyperparameters given in Sec.\ \ref{Sec:PINNloss}. Not documented here are the loss curves, which we have monitored prior to performing the experiments for this paper. In order to reach saturation in the training loss curve, we decided to run the optimisation for the simple harmonic oscillator for 100,000 epochs. The other differential equations are trained for 50,000 epochs. This difference can be related to the oscillatory characteristic that we will later see in the results.    

For time-dependent ODEs (radioactive decay and simple harmonic oscillator), the PINN model is trained using $100$, $200$, and $400$ training points. The trained model is evaluated using 500 equidistant points to ensure that points not included in the training set are also examined. Likewise, for two-dimensional PDEs (Laplace and Poisson), training grids of $20 \times 20$, $40 \times 40$, and $80 \times 80$ interior points are used to train the model. In addition, the edge of the square domain is discretised into $30$ boundary points. This setup ensures that the network is trained with both interior and boundary information. For evaluation purposes, a grid with $100\times100$ equidistantly spaced points is used. The training points are generated using all five training point distribution strategies discussed in the section \ref{training_point_generation}.  

For the subsequent experiments, we first utilise a deterministic fixed-seed (random seed 42) weight initialisation for numerical reproducibility and several random initialisations in the second part. This is achieved with an implementation of the PINN framework in Python using tensorflow \cite{tensorflow2015-whitepaper}, sklearn \cite{scikit-learn}, scipy \cite{2020SciPy-NMeth} and other libraries. 

Let us recall that the purpose of this study is not to find state-of-the-art results, but to investigate the characteristics of various training data distribution strategies. Therefore, we do not challenge our results, and we are completely aware that standard methods from numerical mathematics like, e.g., Runge-Kutta 4 for initial value problems, will most likely outperform our results.

\section{Experiments and Results}

Our experimental design provides a reference point to see the effect of grid sizes and network depths over various training point distribution strategies for solving differential equations. The results with visualisations and conclusions for each differential equation are presented in the following sections.

\subsection{Fixed-seed weight initialisation}

Results in this section use a deterministic (random seed 42) initialisation for the neural network weights. The weight initialisation has a direct impact on the final prediction, as these values dictate the starting position for the optimisation in the weight space. Therefore, the initial set of weights can be very close to the global minimum or far away from a suitable minimum. Usually, first-order optimisation methods like Adam only find local minima \cite{algoritmy,kingma2014adam}. However, for scientific purposes, results should be reproducible and this motivates us to deliver our findings based on a fixed-seed. In the second part, the variance across the optimisations with different random seeds will be discussed.

\subsubsection{Radioactive Decay}
\label{Sec:fixedSeedRadio}
This section presents the mean absolute error (MAE) of the PINN-predicted solution for the radioactive decay problem, evaluated against the analytical solution. For this problem, the initial condition is set to  $x_{0}=100$ and the model was trained in the time domain of $[0,20]$. The solution is shown in Fig.\ \ref{fig:PINNvsExact_Radio}.

\begin{figure}[!h]
\centering
\includegraphics[width=0.75\linewidth]{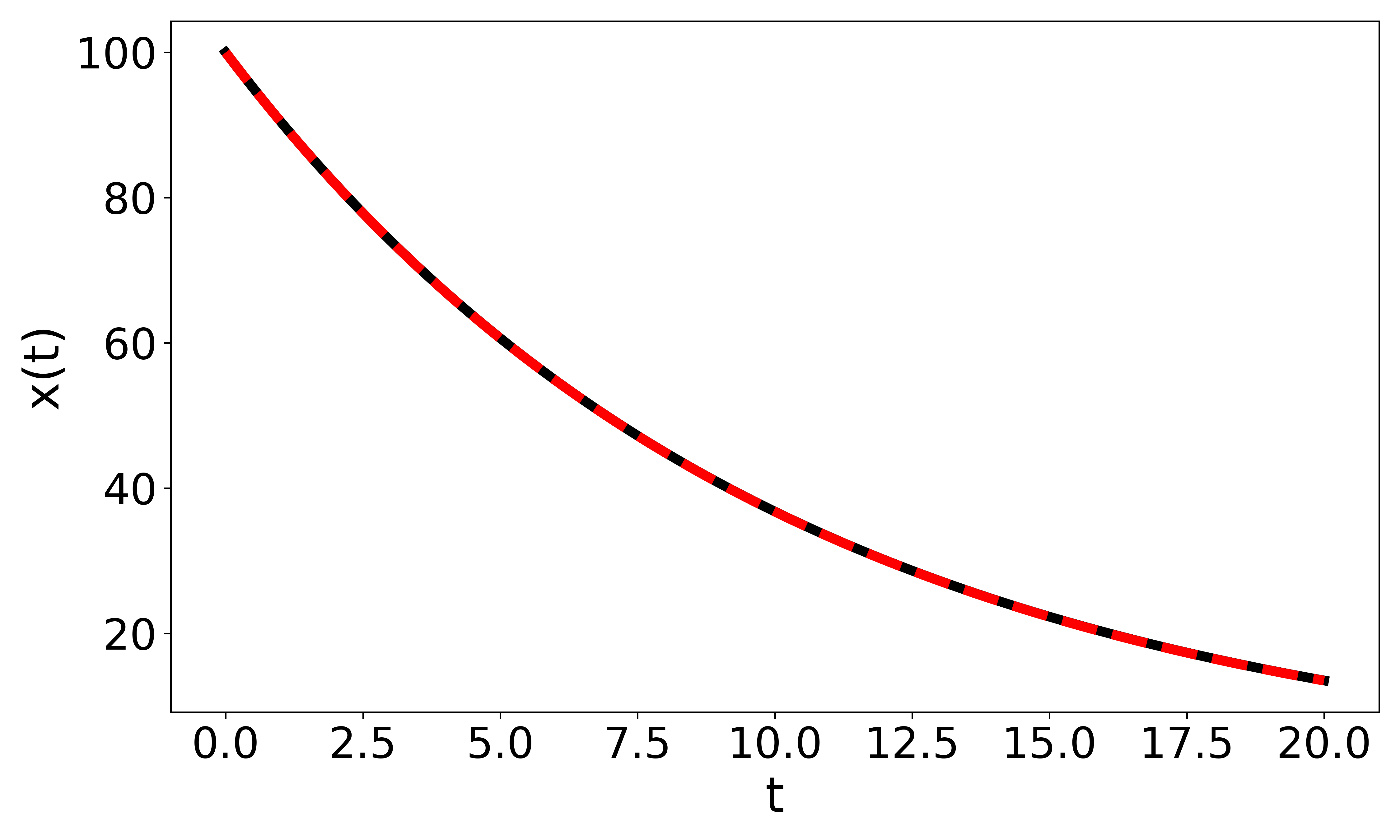}
\caption{Visualisation of the exact vs. PINN solution for the radioactive decay (Sec.\ \ref{Sec:fixedSeedRadio}). (black) exact solution, (red) PINN solution. The PINN solution is from the one-layer FNN trained with 400 equidistant points.}
\label{fig:PINNvsExact_Radio}
\end{figure}

\begin{figure}[!h]
\centering
\includegraphics[width=0.75\linewidth]{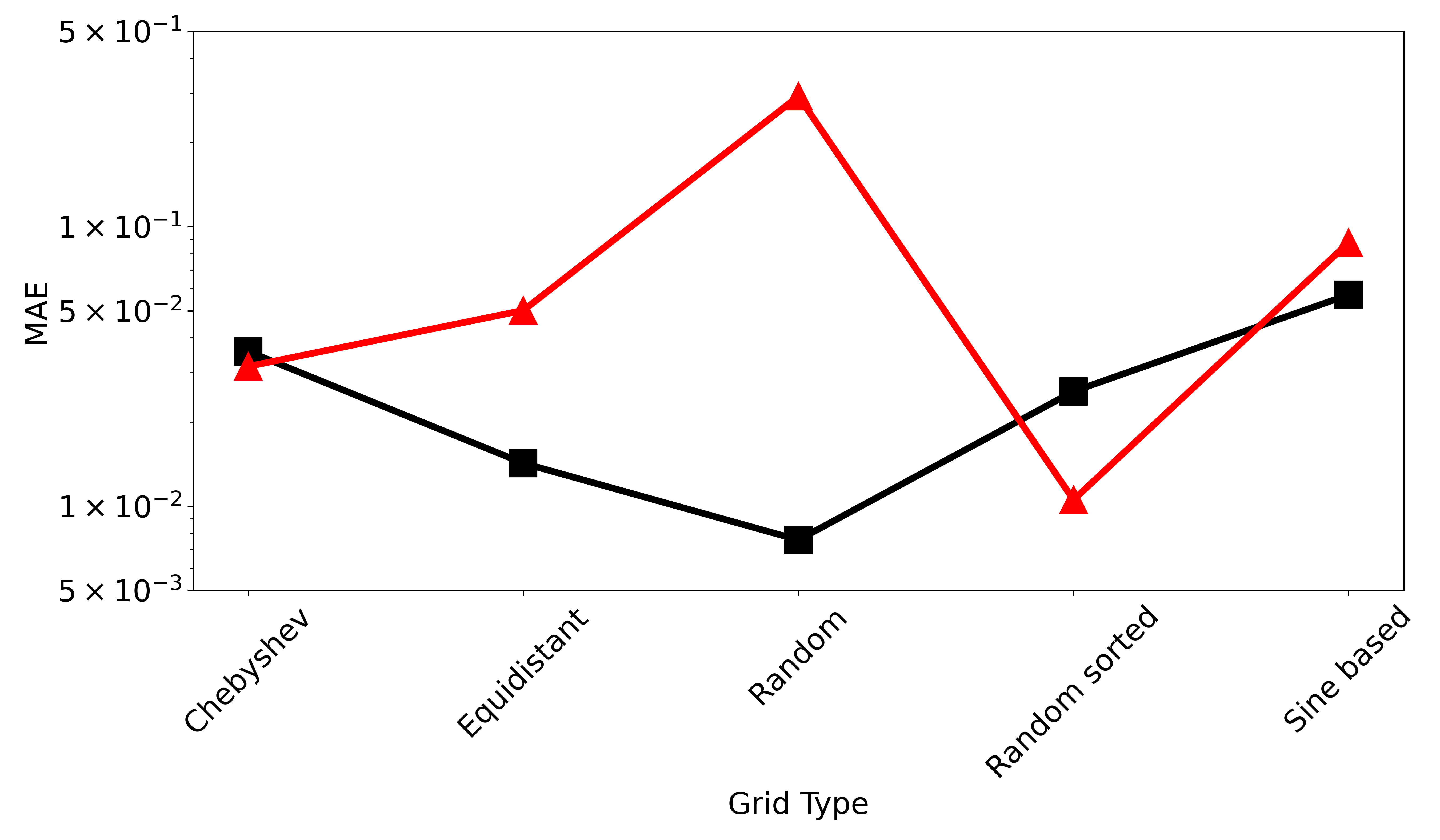}
\caption{Mean absolute error (MAE) results comparing one vs. two hidden layers for radioactive decay with fixed-seed weight initialisation (Sec.\ \ref{Sec:fixedSeedRadio}). (black/cube) one hidden layer with 100 neurons, (red/triangle) two hidden layer with 50 neurons each.}
\label{fig:1vs2layerRadioMAE400grid}
\end{figure}

The plots in Fig.\ \ref{fig:1vs2layerRadioMAE400grid} show results for five training point distribution strategies (400 grid points each) using FNNs with one hidden layer (black/cube) and two hidden layers (red). While the Chebyshev distribution has the smallest difference between the FNNs, the random grid shows a large gap. The overall best performance here is achieved by random grid/one hidden layer and random sorted/two hidden layers. This result is in particular interesting, since the random grid distribution (both uniform and sorted) is also seed-dependent and can be completely different for another seed. Since the radioactive decay, as seen in Fig.\ \ref{fig:PINNvsExact_Radio} shows a rapid decay and converges towards zero, we expected the Chebyshev and Sine-based distributions to perform best. This is because of their characteristic of having more training points at the boundaries and therefore, in our assumption, having more potential to capture the rapid decaying nature. The general trend for the radioactive decay tends to prefer a single hidden layer FNN with a sorted random grid distribution for a fairly high grid size.

\begin{figure}
\centering
\subcaptionbox{One hidden layer with 100 neurons}{\includegraphics[scale=0.4]{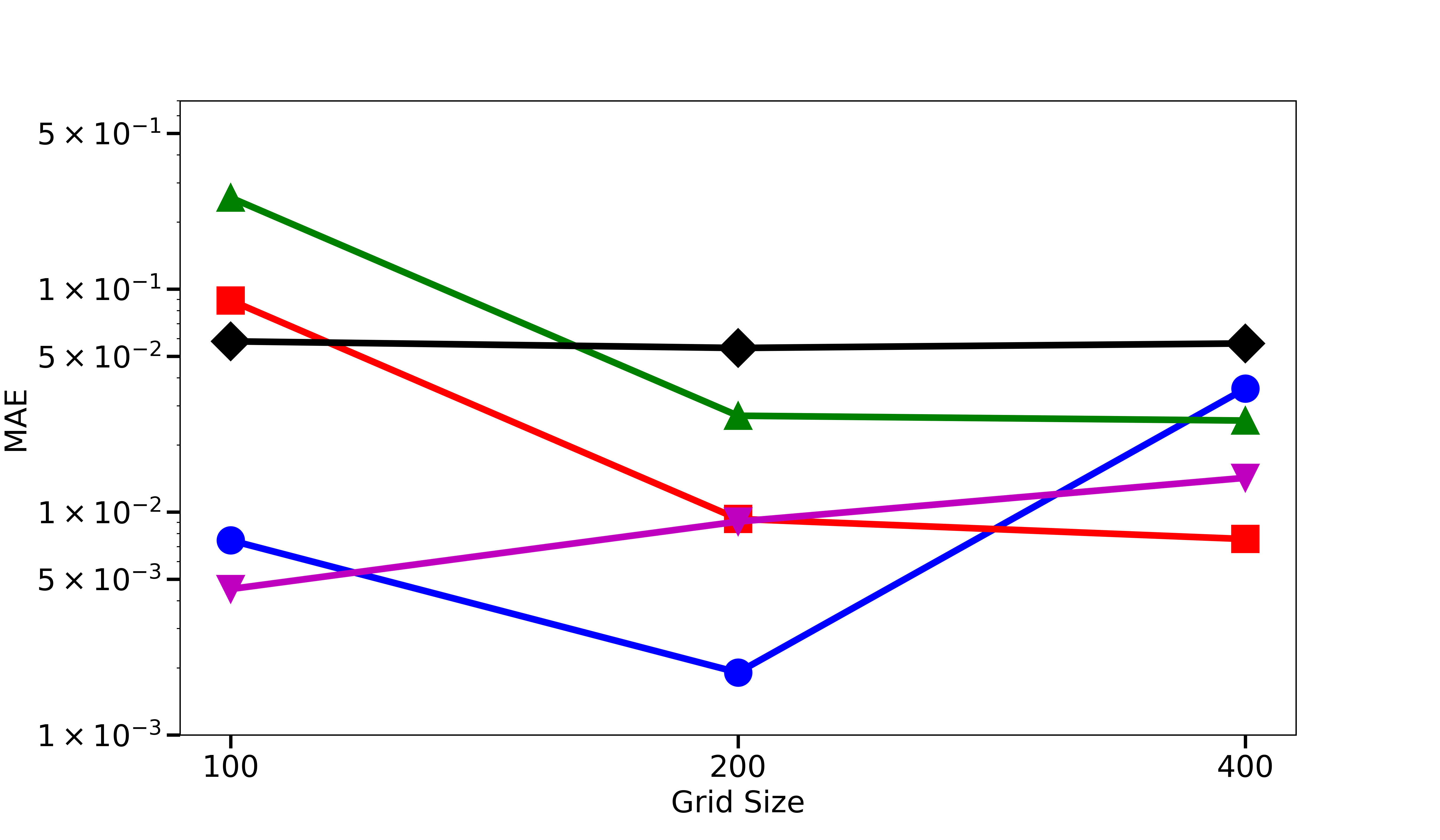}} \\
\subcaptionbox{Two hidden layers with 50 neurons each}{\includegraphics[scale=0.4]{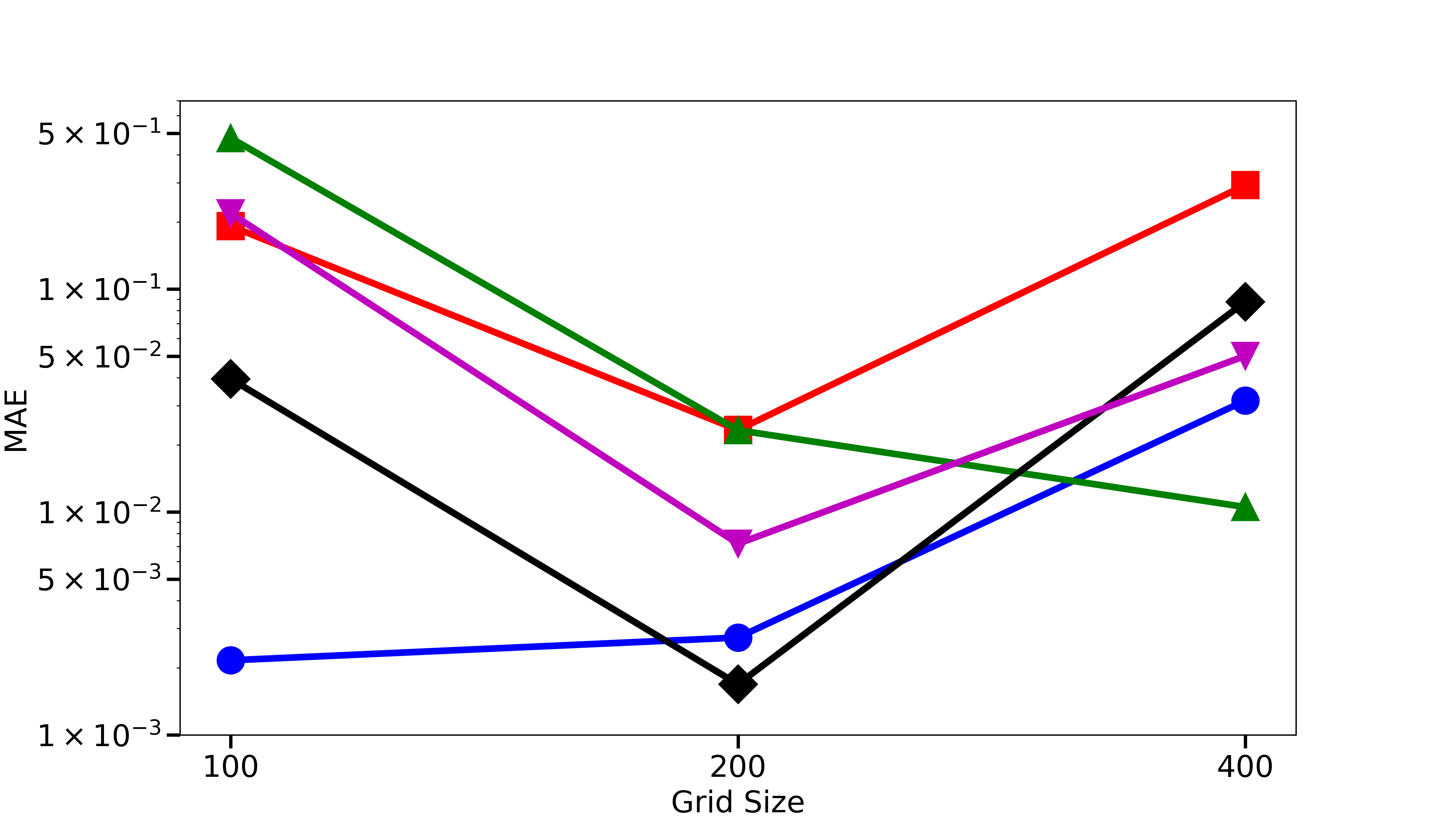}}
\caption{Mean absolute error (MAE) results comparing different training point distributions for radioactive decay with fixed-seed weight initialisation (Sec.\ \ref{Sec:fixedSeedRadio}). (blue/point) Chebychev, (red/cube) Random, (green/triangle up) Random sorted, (black/diamond) Sine based, (purple/triangle down) Equidistant.}
\label{fig:1vs2layerRadioMAEdiffGrid}
\end{figure}

Challenging the findings for Fig.\ \ref{fig:1vs2layerRadioMAE400grid}, different grid sizes are taken into account in Fig.\ \ref{fig:1vs2layerRadioMAEdiffGrid}. The overall best results are achieved by using a combination of one hidden layer with Chebyshev grid distribution and 200 grid points, as well as two hidden layer with a random grid and 200 grid points. Having 100 or 400 grid points leads to worse results for the previously mentioned best combinations. In general, a more dense grid tends to significantly decrease the accuracy when used with a two hidden layer FNN. Only the random sorted grid achieves a performance increase. For the single layer FNN, there is no clear statement available for the behaviour of the different strategies. 

The random sorted grid shows the almost exact trend as the random grid for one hidden layer, but with lesser accuracy. For two hidden layer and 200 grid points, both distributions have almost the same accuracy, which is then split into a significant worse MAE for random and a significant better MAE for the random sorted grid and 400 points. 

As already stated above, the expected the Chebyshev grid to show the best performance and it turns out that this is partially true for one hidden layer with 200 grid points and for two hidden layer with 100 grid points. This is a clear confirmation of the various number of parameters that have an impact on the PINN performance.



\subsubsection{Simple harmonic oscillator}
\label{Sec:fixedSeedSHO}
\begin{figure}[!h]
\centering
\includegraphics[width=0.75\linewidth]{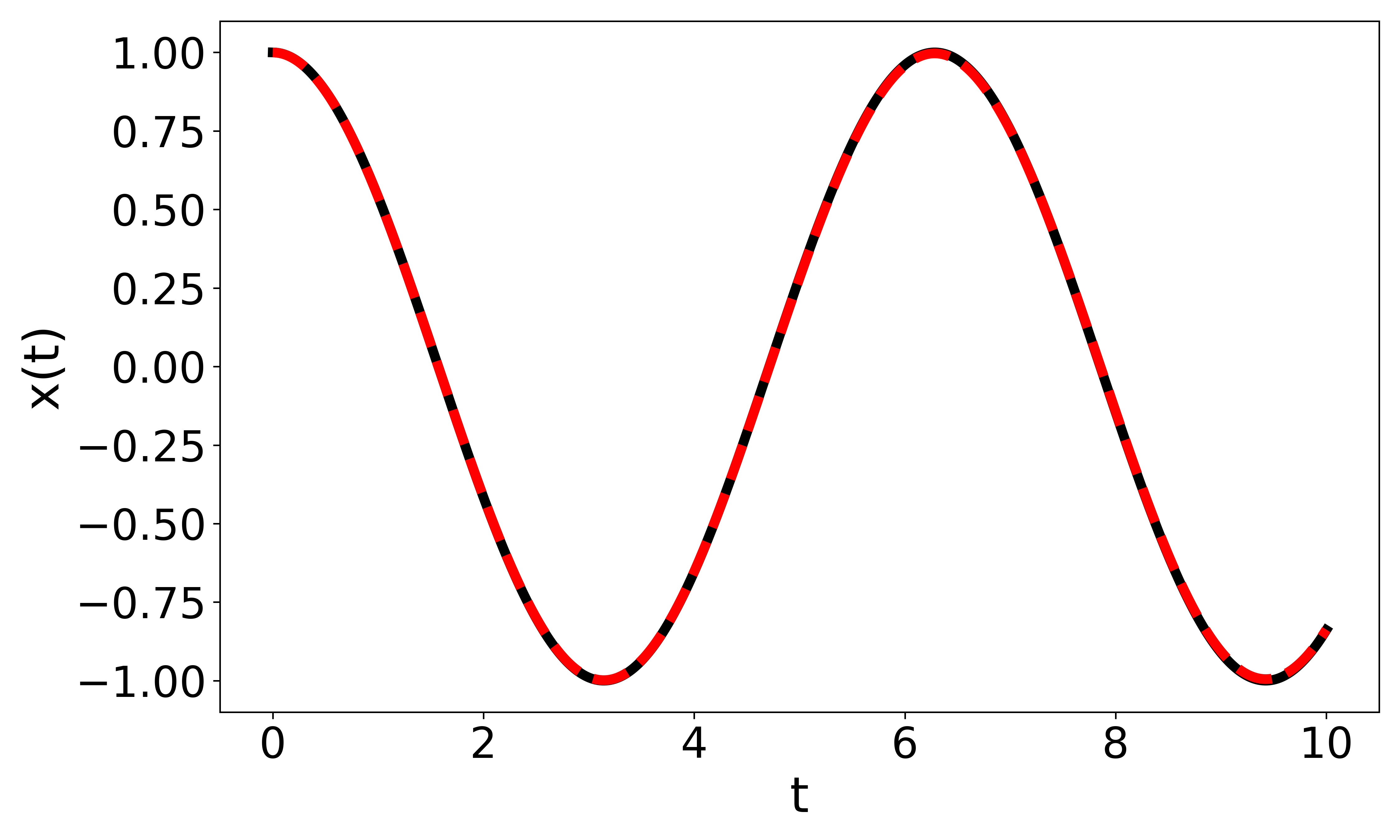}
\caption{Visualisation of the exact vs. PINN solution for the simple harmonic oscillator (Sec.\ \ref{Sec:fixedSeedSHO}). (black) exact solution, (red) PINN solution. The PINN solution is from the one-layer FNN trained with 400 equidistant points.}
\label{fig:PINNvsExact_SHO}
\end{figure}

This section presents the mean absolute error (MAE) of the PINN-predicted solution for the simple harmonic oscillator problem, evaluated against the analytical solution. For this problem, the initial displacement $x_0$, and initial velocity $v_0$ are set to be $1$ and $0$, respectively. Angular velocity is set to be $\omega=1$, and the model was trained in the time domain of $[0,10]$. The solution is shown in Fig.\ \ref{fig:PINNvsExact_SHO}.

\begin{figure}[!h]
\centering
\includegraphics[width=0.75\linewidth]{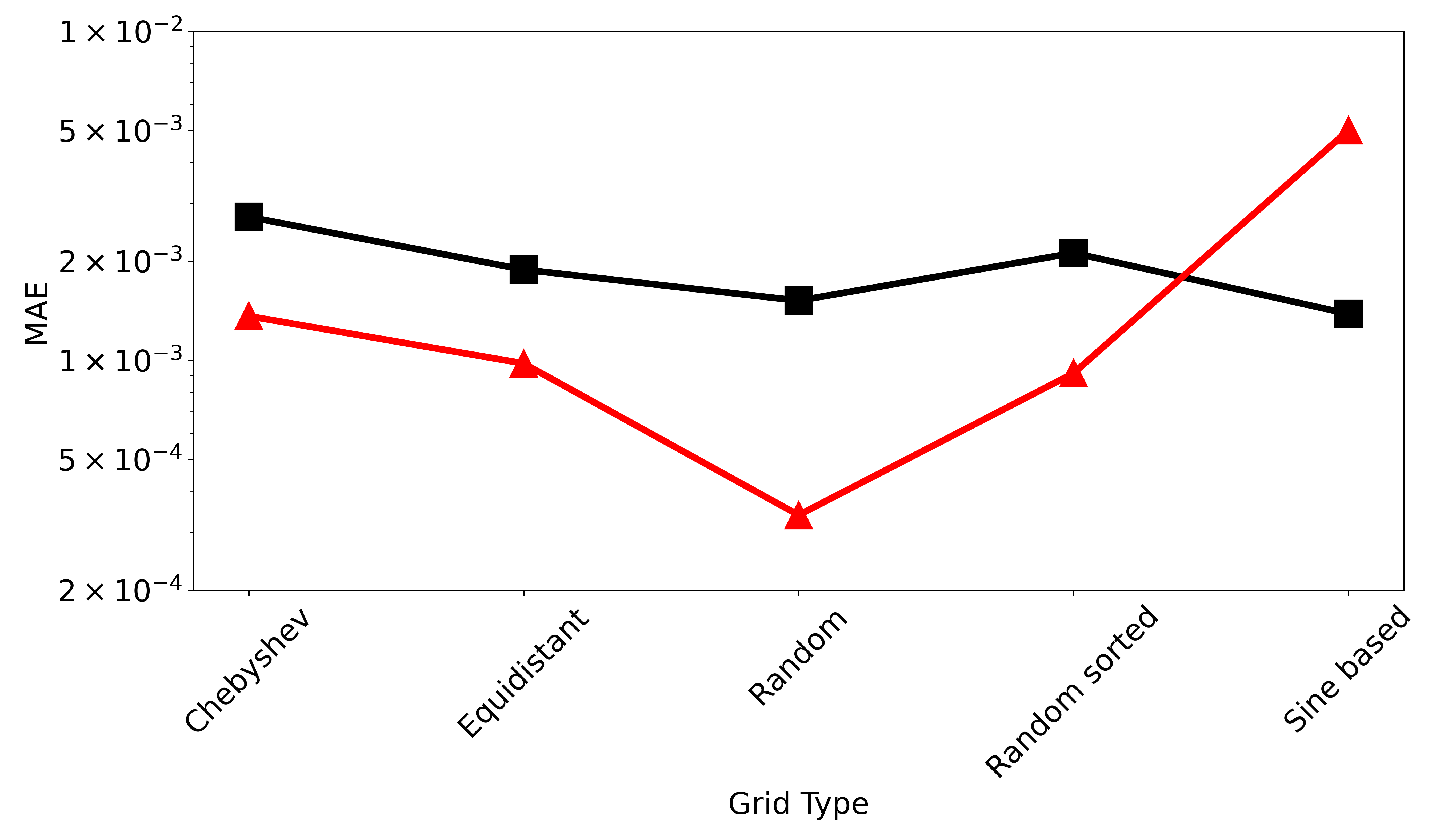}
\caption{Mean absolute error (MAE) results comparing one vs. two hidden layers for simple harmonic oscillator with fixed-seed weight initialisation (Sec.\ \ref{Sec:fixedSeedSHO}). (black/cube) one hidden layer with 100 neurons, (red/triangle) two hidden layer with 50 neurons each.}
\label{fig:fixedSeedSHO_1_vs_2_layer}
\end{figure}
The plot in Fig \ref{fig:fixedSeedSHO_1_vs_2_layer} shows the results for five training point distribution strategies (400 grid points each) using FNNs with one hidden layer (black/cube) and two hidden layers (red/triangle). When adding an extra hidden layer and distributing the number of neurons equally, e.g., 50 neurons each, the MAE value has improved for almost every grid type. Overall, the best performance here is obtained by the sine-grid for a two-layer FNN. Here, the assumption was that a well-clustered point distribution can capture the oscillatory nature of the simple harmonic oscillator. This seems true for the sine-based grid when used to train a two-layer FNN.

However, the clustering and the solution nature may not exactly provide the necessary concentration for an arbitrary number of grid points. Considering the 200 grid points, the sine-based distribution shows the best accuracy, while the Chebyshev distribution provides the worst accuracy. This is interesting behaviour since both strategies share a more dense clustering at the boundaries. However, the nature of the sine-based distribution seems to be really important for matching with the solution nature of the simple harmonic oscillator. Nonetheless, this only holds true when training points are carefully distributed.



\begin{figure}
\centering
\subcaptionbox{One hidden layer with 100 neurons}{\includegraphics[scale=0.4]{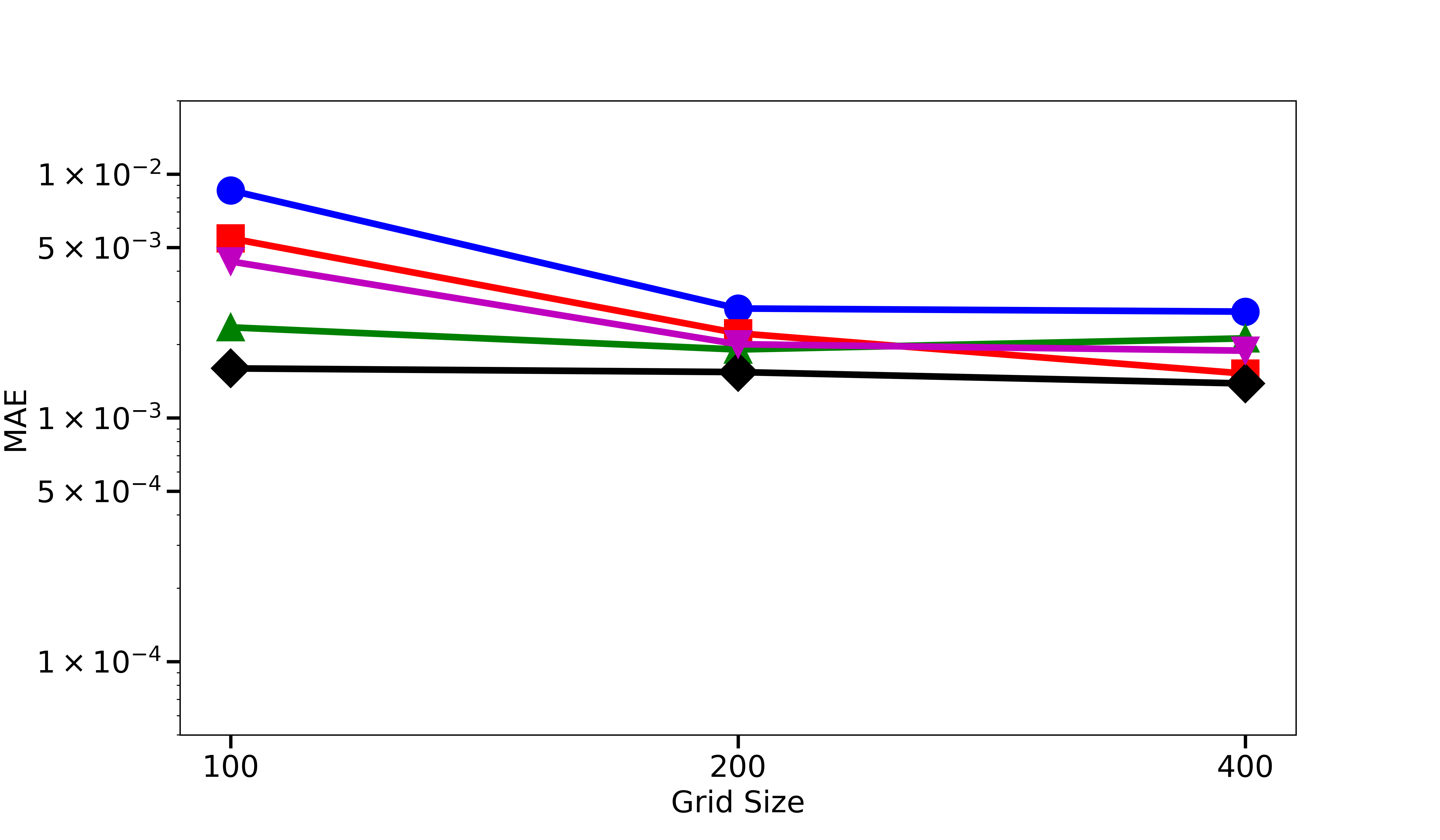}} \\
\subcaptionbox{Two hidden layers with 50 neurons each}{\includegraphics[scale=0.4]{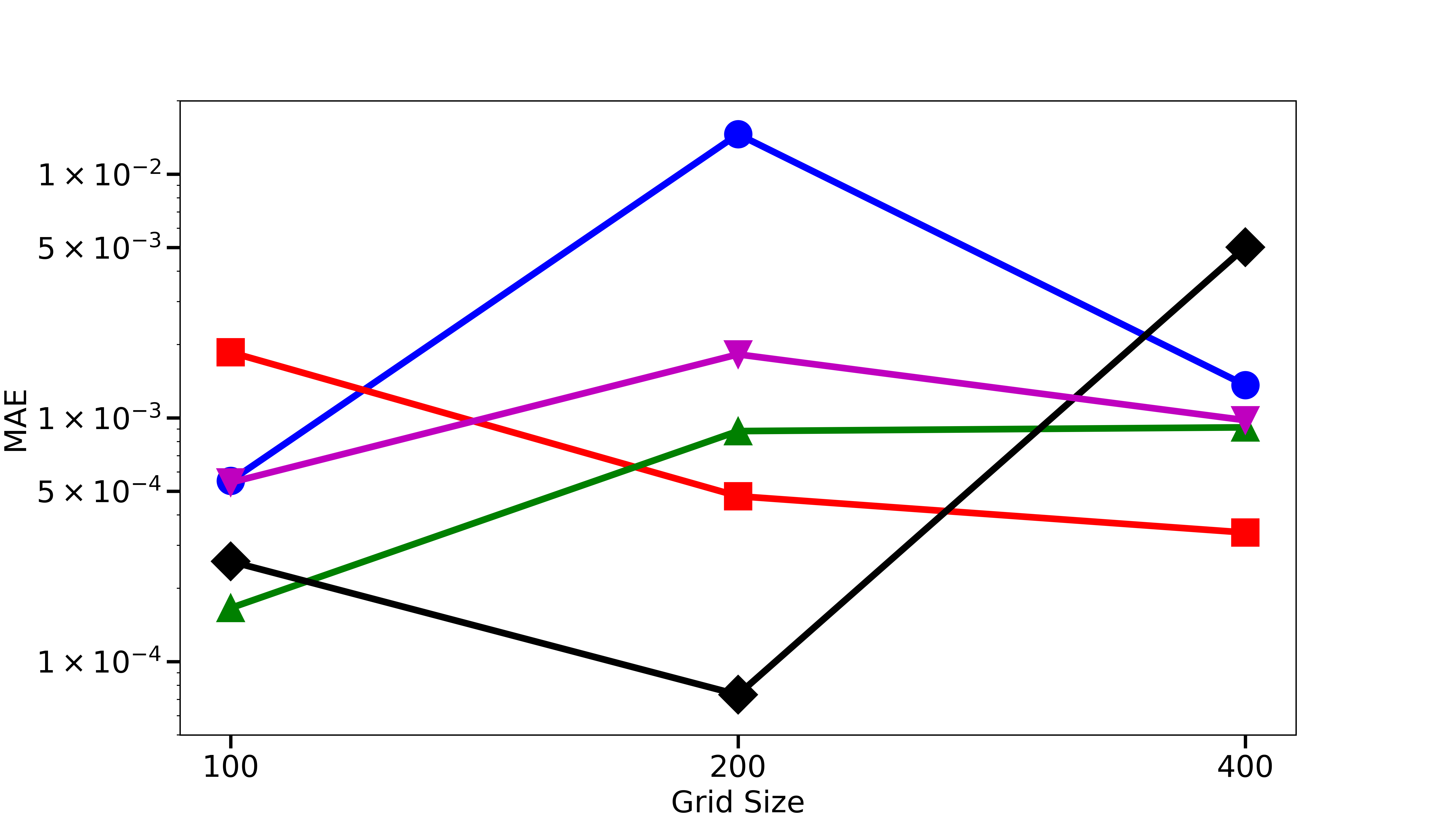}}
\caption{Mean absolute error (MAE) results comparing different training point distributions for simple harmonic oscillator with fixed-seed weight initialisation (Sec.\ \ref{Sec:fixedSeedSHO}). (blue/point) Chebychev, (red/cube) Random, (green/triangle up) Random sorted, (black/diamond) Sine based, (purple/triangle down) Equidistant.}
\label{fig: MAEvsGridsize_SHO}
\end{figure}

\subsubsection{Laplace equation}
\label{Sec:fixedSeedLaplace}

\begin{figure}[!h]
\centering
\includegraphics[width=1\linewidth]{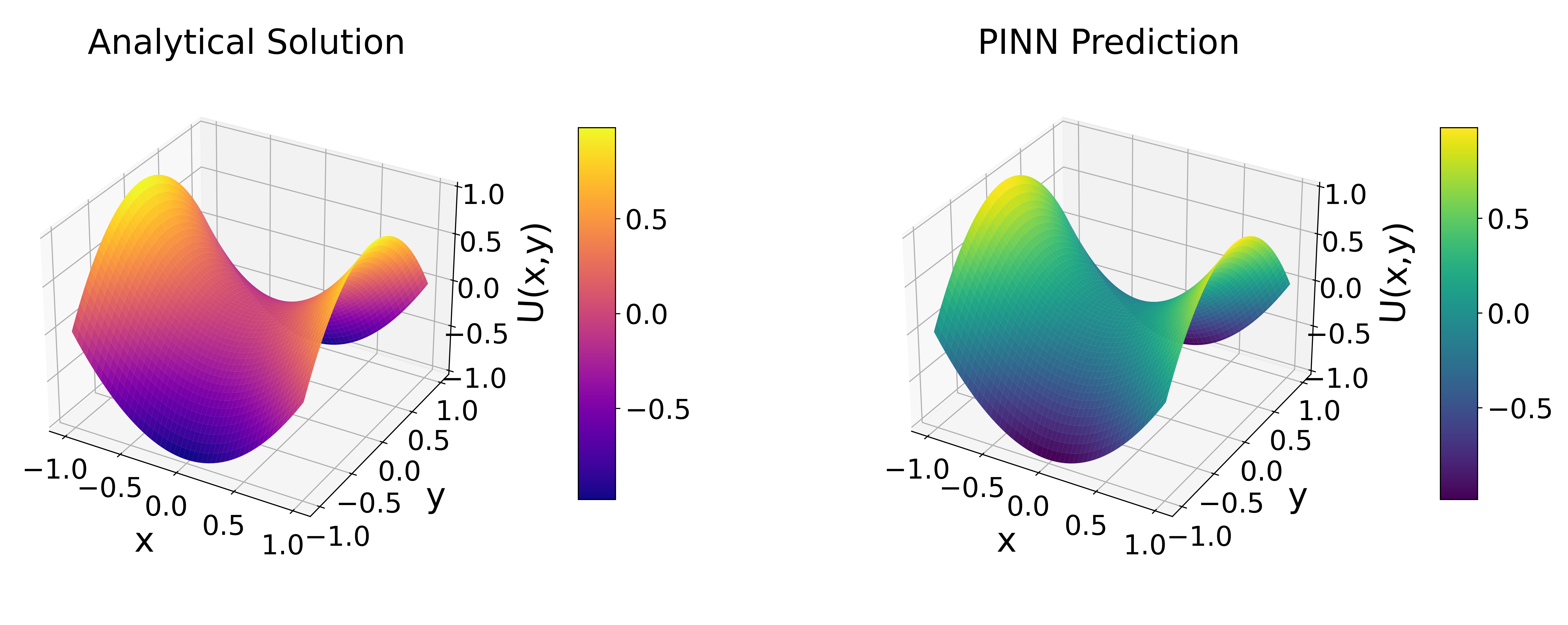}
\caption{Visualisation of the exact vs. PINN solution for the Laplace equation (Sec.\ \ref{Sec:fixedSeedLaplace}). (left) exact solution, (right) PINN solution. The PINN solution is from the one-layer FNN trained with $80\times80$ equidistant points.}
\label{fig:PINNvsExact_Radio}
\end{figure}
The two-dimensional Laplace equation is solved using the same FNN that has already been discussed in section \ref{Sec:fixedSeedRadio} and \ref{Sec:RandomSeedSHO}. The only difference to the model is that it now has a two-dimensional spatial input $(x,y)$. In this section, we present the comparison of the accuracy of this model in predicting the solution to the Laplace equation for various grid distribution strategies and grid sizes for both one and two-layer FNN architectures. 

\begin{figure}[!h]
\centering
\includegraphics[width=0.75\linewidth]{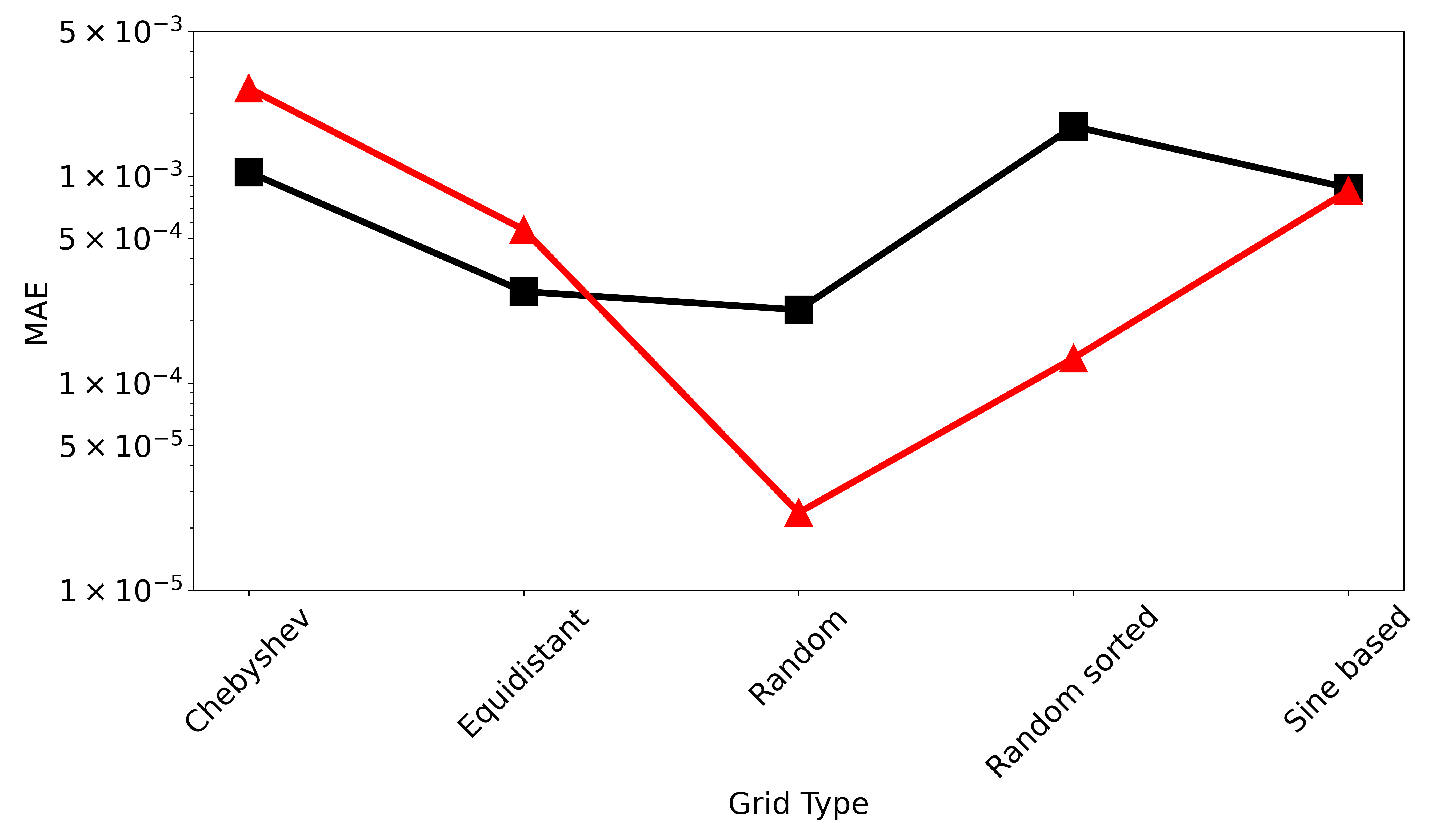}
\caption{Mean absolute error (MAE) results comparing one vs. two hidden layers for Laplace equation with fixed-seed weight initialisation (Sec.\ \ref{Sec:fixedSeedLaplace}). (black/cube) one hidden layer with 100 neurons, (red/triangle) two hidden layer with 50 neurons each.}
\label{fig:1_vs_2_layer_Laplace_80_grid}
\end{figure}

Figure \ref{fig:1_vs_2_layer_Laplace_80_grid} shows the comparison of Laplace PINN accuracy of one and two-layer FNN over various grid types. Both FNNs are trained using $80\times80$ grid size in the square domain of $[-1, 1]^2$. The accuracy of this PINN model has improved for the random grid by adding an extra hidden layer. However, this can only holds for the utilised random (distribution) seed. For a different set of random points, the outcome can be different and cannot be generalised. For structured grids, the accuracy has slightly worsened or remains similar (sine-based). 
\begin{figure}
\centering
\subcaptionbox{One hidden layer with 100 neurons}{\includegraphics[scale=0.4]{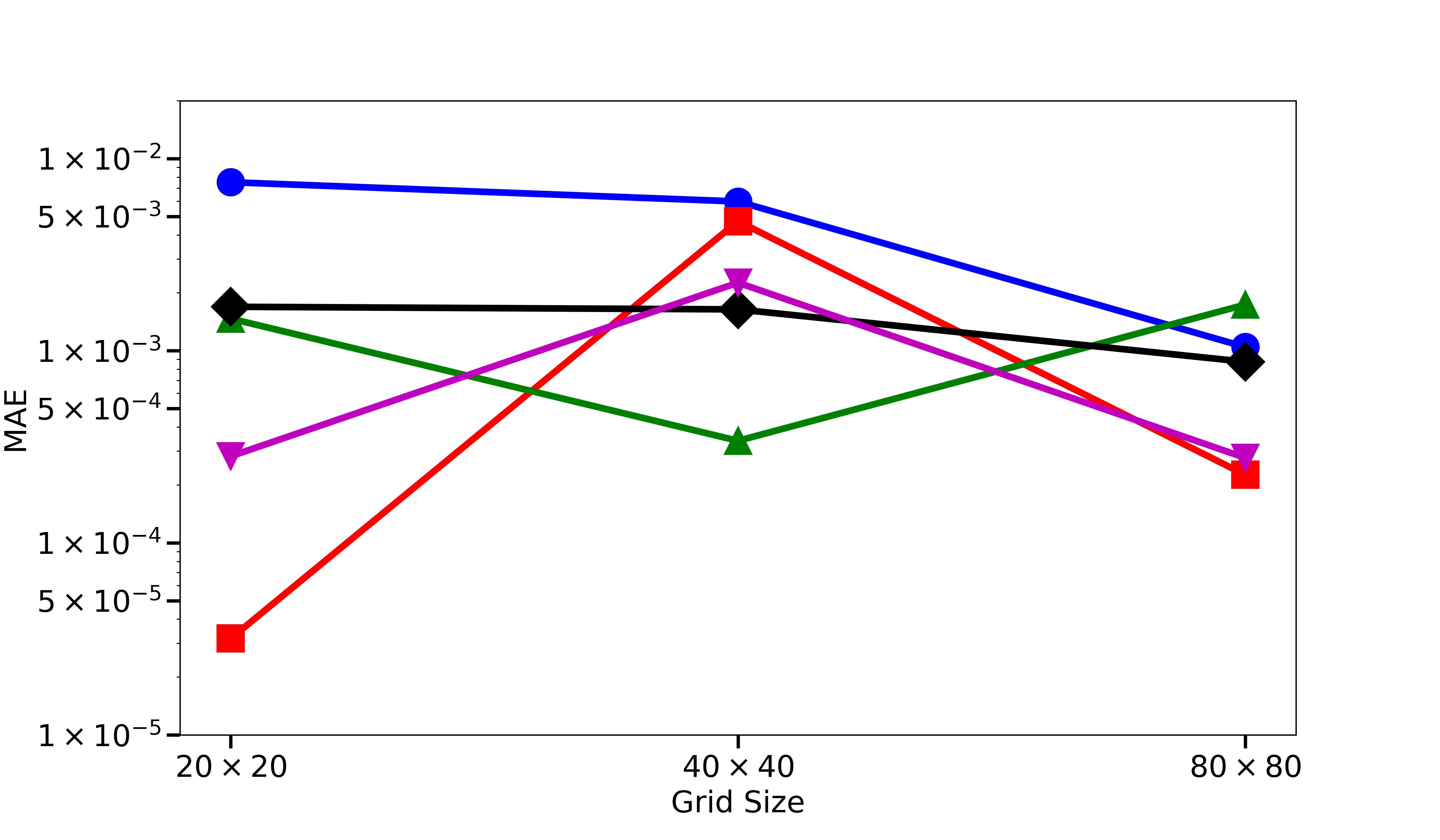}} \\
\subcaptionbox{Two hidden layers with 50 neurons each}{\includegraphics[scale=0.4]{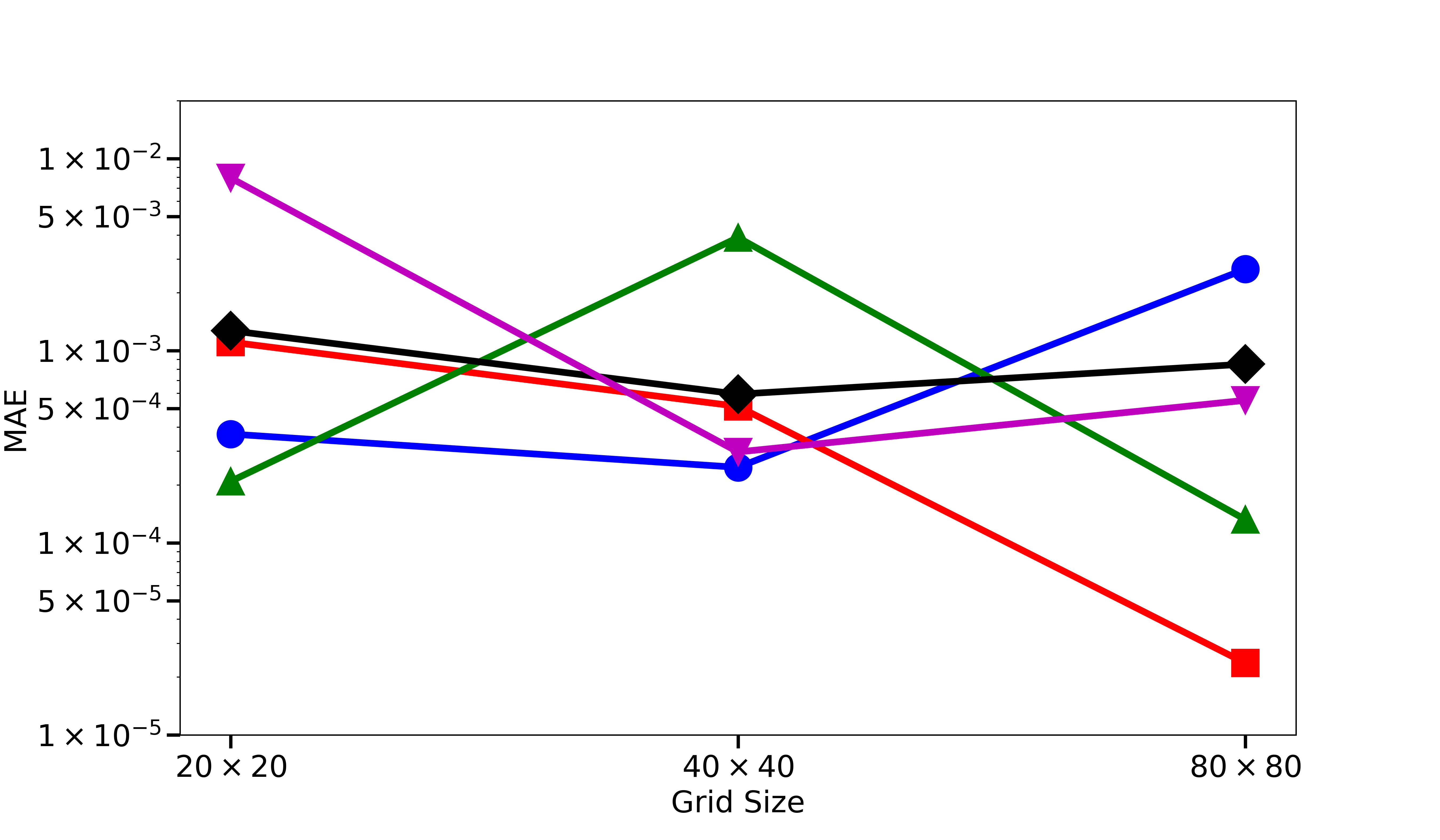}}
\caption{Mean absolute error (MAE) results comparing different training point distributions for Laplace equation with fixed-seed weight initialisation (Sec.\ \ref{Sec:fixedSeedLaplace}). (blue/point) Chebychev, (red/cube) Random, (green/triangle up) Random sorted, (black/diamond) Sine based, (purple/triangle down) Equidistant.}
\label{fig: Laplace_side_by_side}
\end{figure}
To further evaluate the performance of this PINN model, the model is also trained using $20\times20$ and $40\times40$ grid points, in addition to the $80\times 80$ grid. The respective plots for one and two-layer FNNs showing variation in MAE over grid sizes for all grid types are visualised in Fig. \ref{fig: Laplace_side_by_side}. On both plots (a) and (b) in Figure \ref{fig: Laplace_side_by_side}, one can see the non-monotonic trend in MAE progression over the grid sizes. However, clustered grids (Chebyshev and sine-based) have shown a decrease in MAE as the grid size increases from $20\times20$ to $80\times80$ for one-layer FNN.   
For a two-layer FNN, a similar kind of improvement in accuracy is seen for the random grid. The overall best performance is given by the random grid, for a one-layer FNN at $20\times20$ and a two-layer FNN at $80\times80$ grid size. The Chebyshev grid, which has the worst accuracy for all grid sizes for a one-layer FNN, has performed the best among all grid types for $40\times40$ grid sizes for a two-layer architecture. For the sine-based grid, the accuracy has improved on grid densification for both FNNs; however, the accuracy has slightly worsened when the grid size is increased from $40\times40$ to $80\times80$ for a two-layer FNN.     
\subsubsection{Poisson equation}
\label{Sec:fixedSeedPoisson}
\begin{figure}[!h]
\centering
\includegraphics[width=1\linewidth]{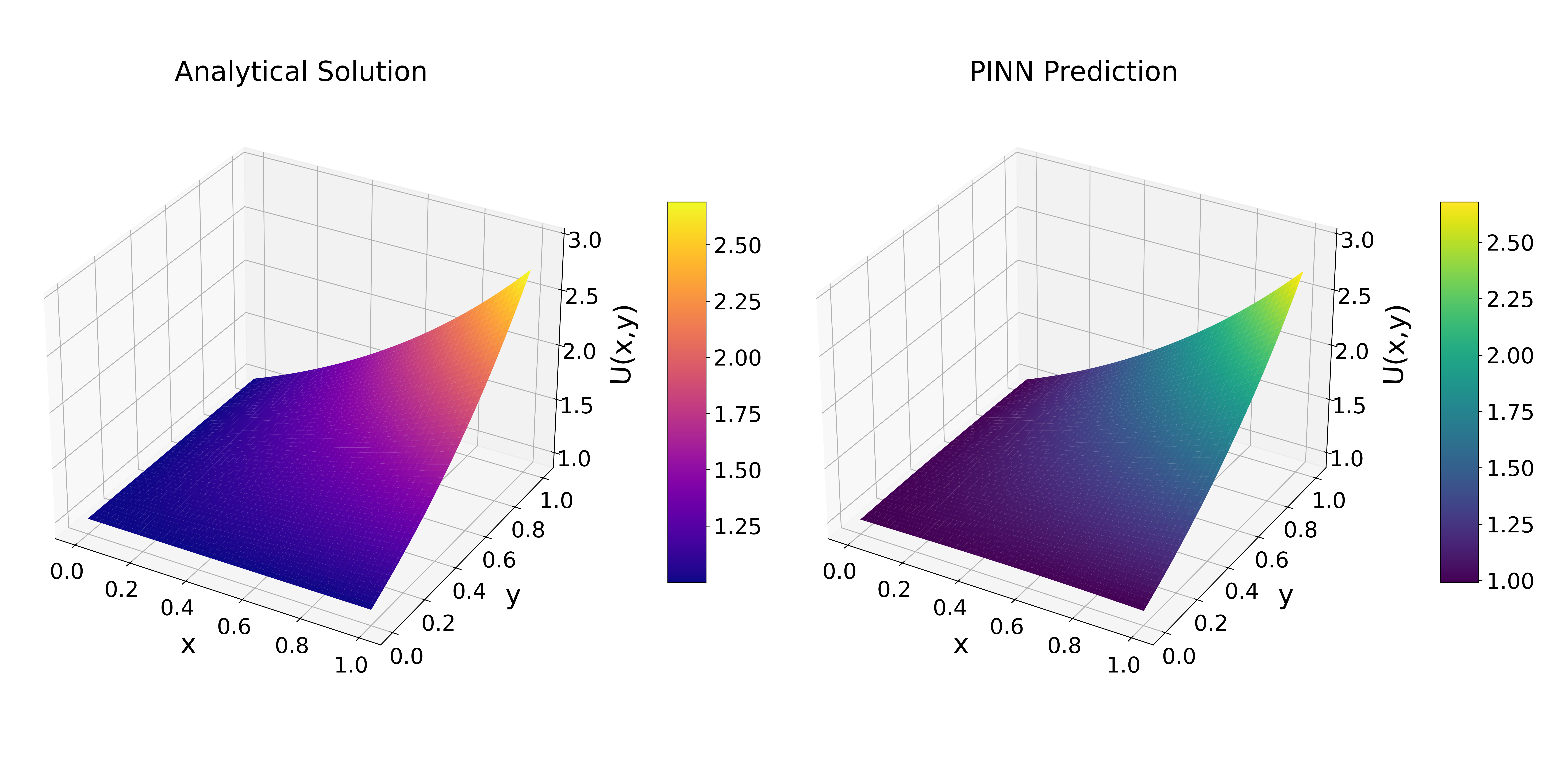}
\caption{Visualisation of the exact vs. PINN solution for the Poisson equation (Sec.\ \ref{Sec:fixedSeedPoisson}). (left) exact solution, (right) PINN solution. The PINN solution is from the one-layer FNN trained with $80\times80$ equidistant points.}
\label{fig:PINNvsExact_Poisson}
\end{figure}

In this section, we present the performance comparison of the PINN model predictions for the Poisson equation. The exact solution and the PINN approximation are visualised in Fig.\ \ref{fig:PINNvsExact_Poisson}.

\begin{figure}[!h]
\centering
\includegraphics[width=0.75\linewidth]{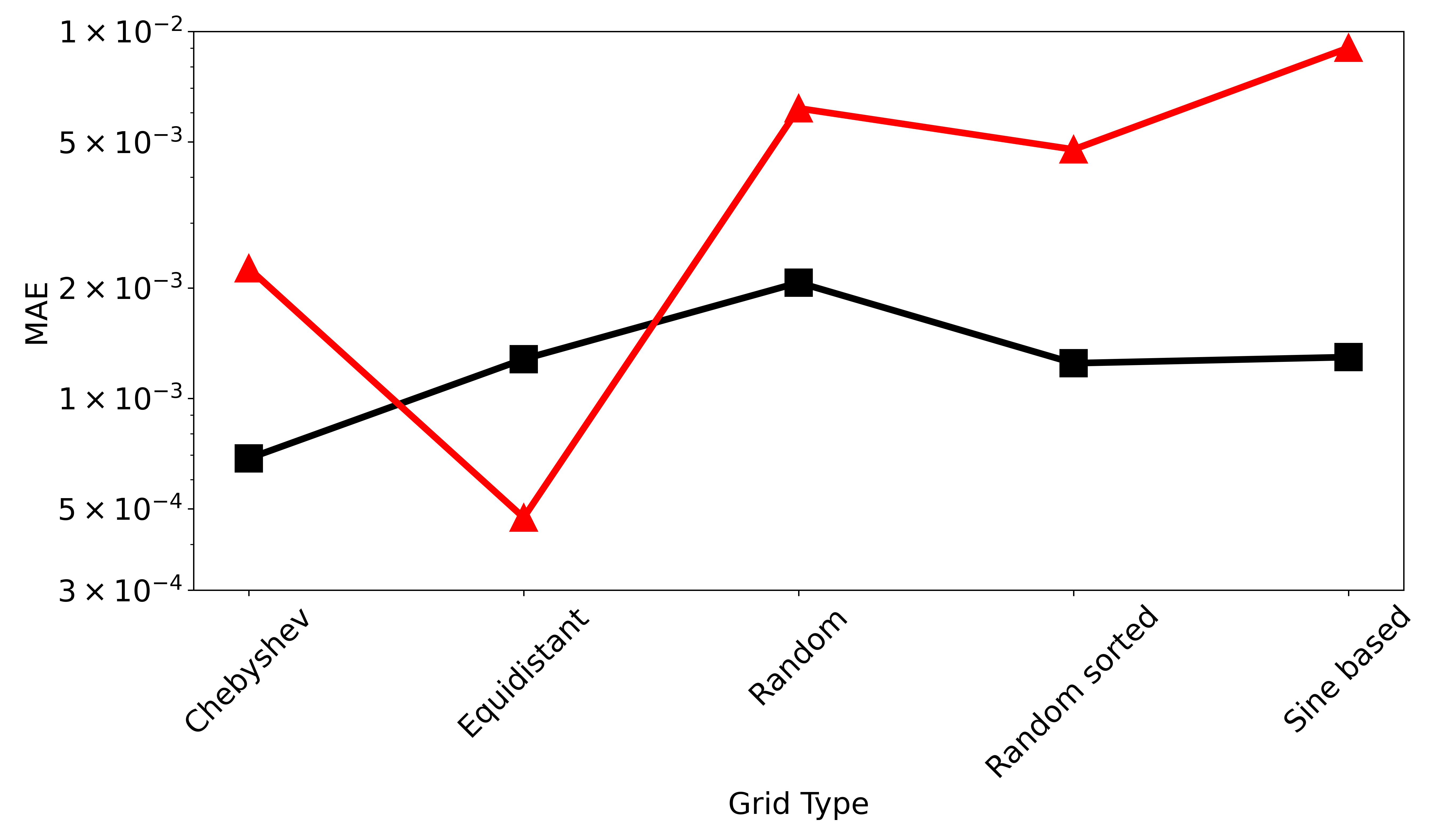}
\caption{Mean absolute error (MAE) results comparing one vs. two hidden layer for Poisson equation with fixed-seed weight initialisation (Sec.\ \ref{Sec:fixedSeedPoisson}). (black/cube) one hidden layer with 100 neurons, (red/triangle) two hidden layers with 50 neurons each.}
\label{fig:1_vs_2_layer_Poisson}
\end{figure}

In Fig. \ref{fig:1_vs_2_layer_Poisson}, we find the performance comparison for one and two-layer FNNs over various grid types when trained using $80\times80$ grid points. The one-layer FNN has a better accuracy than the two-layer FNN, except for the equidistant grid for this particular setup. The reason for this might be that the model has already acquired its best accuracy for those distribution strategies and resolutions, and adding an extra layer significantly increases the number of parameters to be optimised. For this particular PDE, effectively optimising the model becomes more challenging and would require, e.g., more training epochs, for two hidden layers. Both the best and the worst performance came across a two-layer FNN on an equidistant and sine-based grid, respectively.

\begin{figure}
\centering
\subcaptionbox{One hidden layer with 100 neurons}{\includegraphics[scale=0.4]{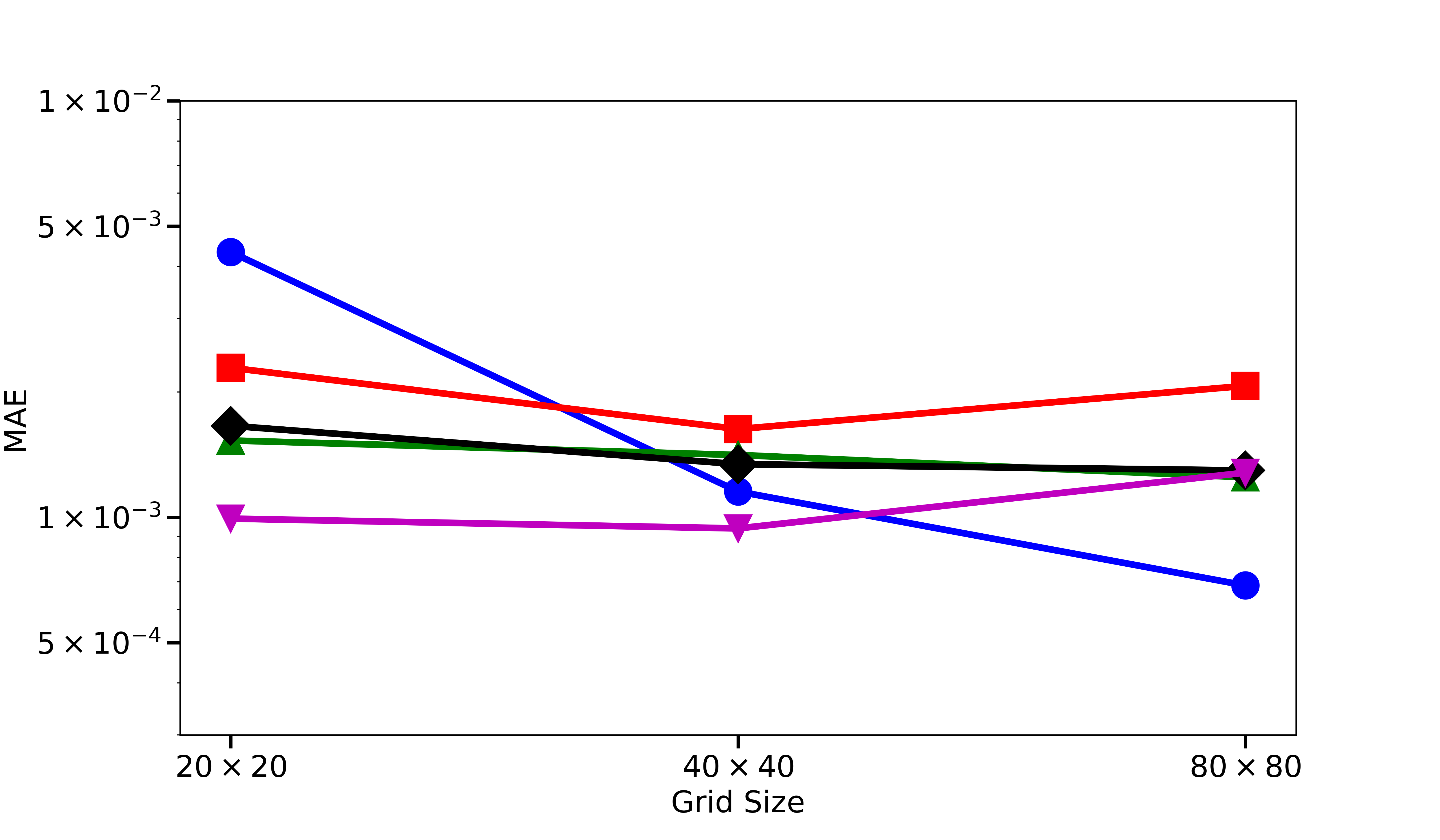}} \\
\subcaptionbox{Two hidden layers with 50 neurons each}{\includegraphics[scale=0.4]{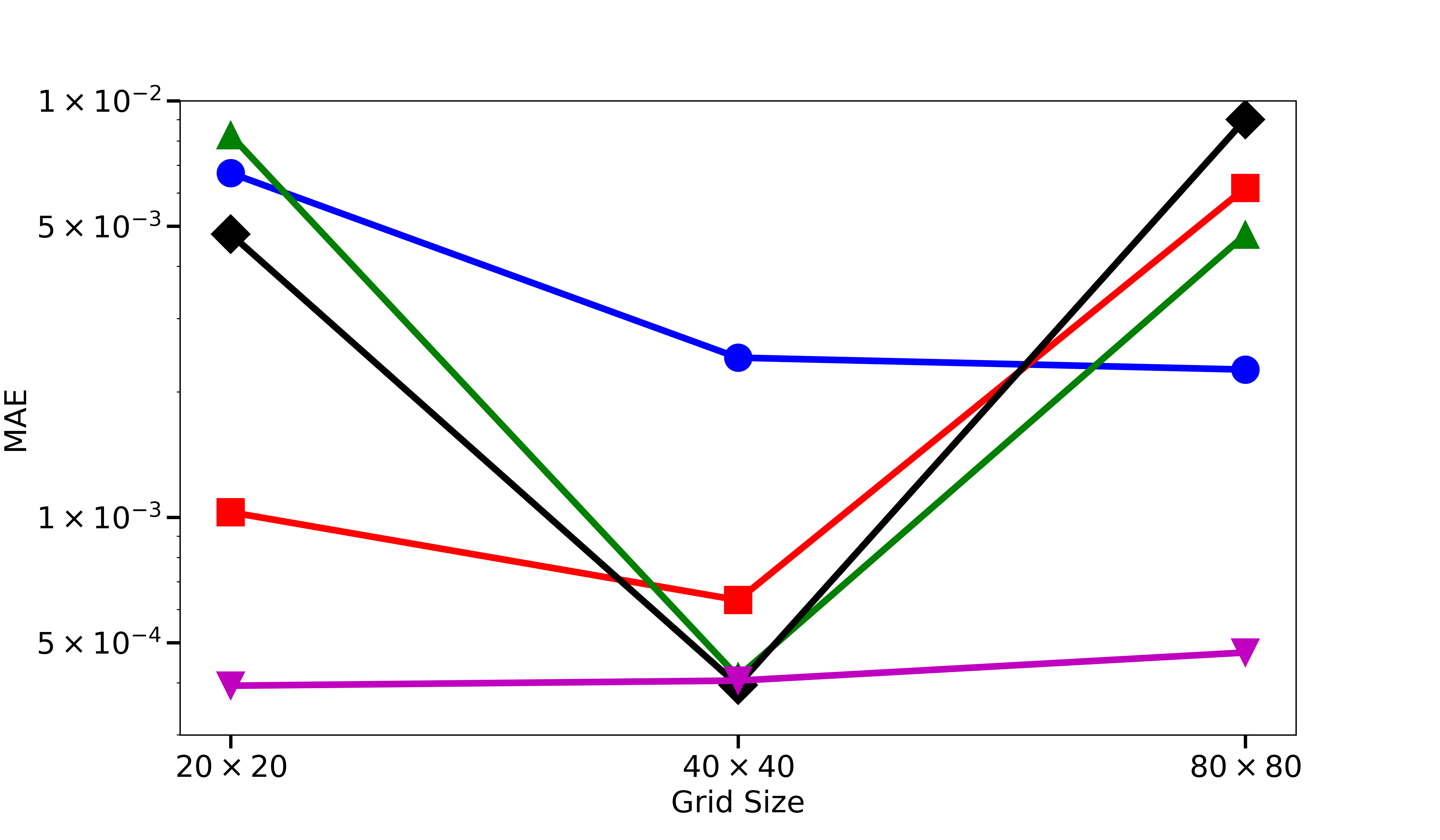}}
\caption{Mean absolute error (MAE) results comparing different training point distributions for Poisson equation with fixed-seed weight initialisation (Sec.\ \ref{Sec:fixedSeedPoisson}). (blue/point) Chebychev, (red/cube) Random, (green/triangle up) Random sorted, (black/diamond) Sine based, (purple/triangle down) Equidistant.}
\label{fig: Poisson_side_by_side}
\end{figure}

To further deep dive into the performance of these PINNs models, both are also trained with $20\times20$ and $40\times40$ grid points with all distribution types. Plots (a) and (b) in Figure \ref{fig: Poisson_side_by_side} show the MAE values against grid sizes for one and two-layer FNNs. All grid types have benefited from increasing grid resolution from $20\times20$ to $40\times40$ for both one and two-layer FNNs. However, a further increase in resolution to $80\times80$ has lowered the accuracy for all grid types and for both FNNs, except for the Chebyshev grid.      

The MAE values for the equidistant grid remain lower than those of other grids, except for one layer FNN at $80\times80$ grid size, where Chebyshev has performed best. The accuracy for the equidistant grid for all sizes has improved with the addition of a hidden layer, indicating that for some grid types, added depth is beneficial for better predicting the solution. However, this is exactly opposite for the Chebyshev grid, for which a two-layer FNN has increased the MAE values for all grid sizes, despite there being a monotonic improvement in accuracy for this grid type for both FNNs with an increase in grid size. Sine-based grid has shown consistent improvement in accuracy with increased grid size for one-layer architecture, whereas for two-layer architecture, it has shown a non-monotonic trend, with a $40\times40$ grid size having the least error value. The random sorted has shown exactly the same trend as that of the sine-based grid for both FNNs. Random grid, for both FNNs, has shown a decrease in MAE with increasing grid size from $20\times20$ to $40\times40$, whereas on further increasing the grid size to $80\times80$, the MAE has increased again.

\subsection{Random weight initialisation}

Considering a single fixed seed for random weight initialisation has benefits when it comes to reproducible characteristics. However, depending on the specific seed, the optimisation may suffer from finding a suboptimal local minimum. Therefore, the initialisation sometimes is, in simple words, luck-based. In this section, we present and discuss results for the simple harmonic oscillator and the Poisson equation for 200 consecutive random seeds.

\subsubsection{Simple harmonic oscillator}
\label{Sec:RandomSeedSHO}

\begin{figure}[!h]
\centering
\includegraphics[width=1\linewidth]{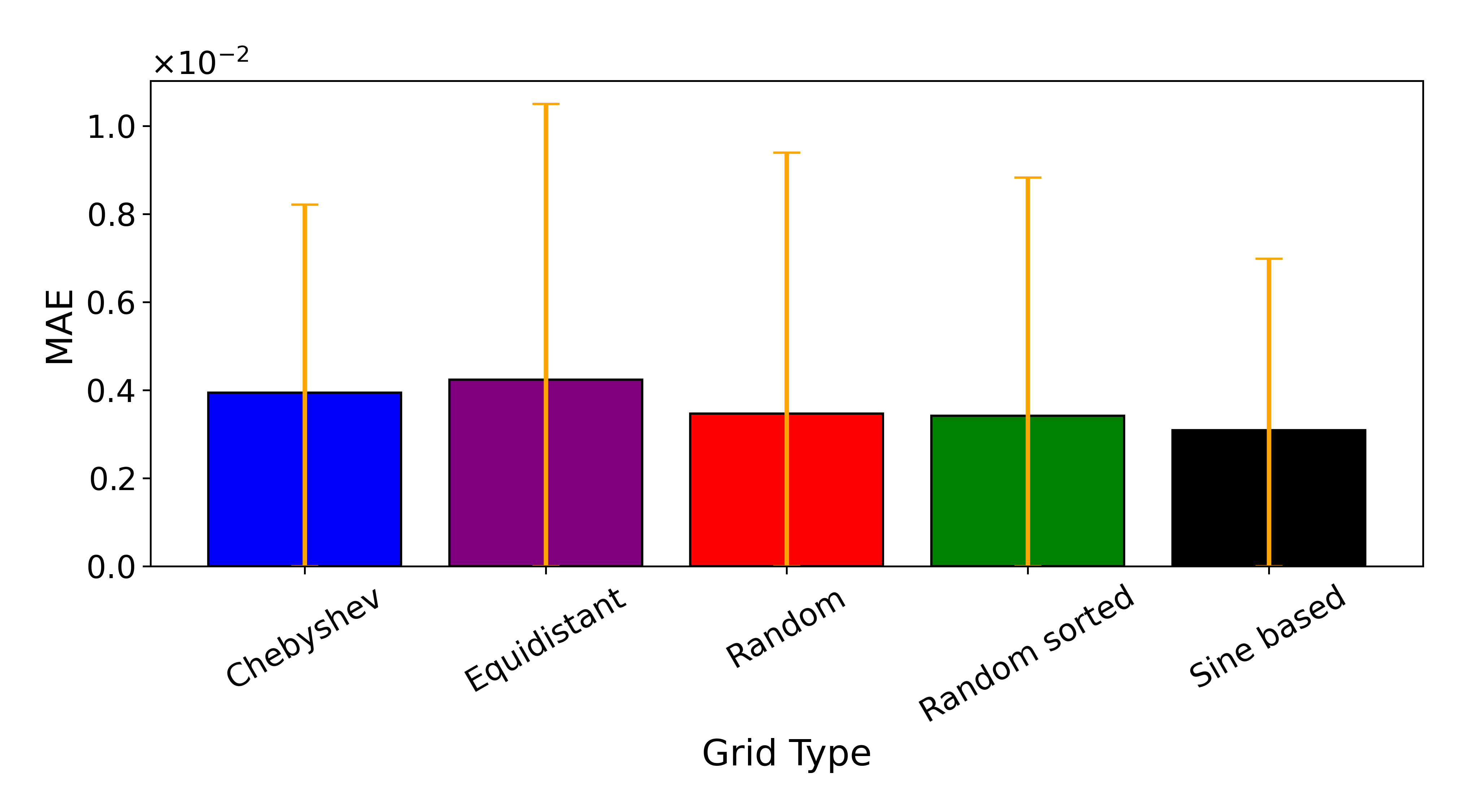}
\caption{Mean absolute errors averaged ($\overline{\text{MAE}}$) over 200 random seeds for the simple harmonic oscillator with a single layer FNN and 400 training points (Sec.\ \ref{Sec:RandomSeedSHO}). The bar colours refer to the $\overline{\text{MAE}}$ of each training point distribution strategy and the standard deviation (SD) is visualised in orange.}
\label{fig:Averaged_MAE_SHO}
\end{figure}

The same simple harmonic oscillator, as discussed in section \ref{Sec:fixedSeedSHO}, is predicted using the PINN model with one hidden layer and 100 neurons. The model is trained using a grid size of 400 points. For each training point distribution, the model is trained for 200 consecutive random seeds, so that a total of 1000 experiments were conducted to provide a performance overview of the PINN on the basis of the averaged mean absolute error for each grid type. 

Tab. \ref{Tab:SHOvalues} lists the loss value at the final epoch, along with their corresponding MAE values for the Chebyshev and Sine-based grid for the particular random seed used to generate the initial weights. This can give readers an idea that, despite having the same inputs and the same FNNs, different initial weight vectors result in different predictions. 
Standard deviations (SD) are represented as vertical yellow error lines in each bar. In all cases, the SD exceeded the corresponding mean MAE, which would result in error bars extending below zero, a scenario that is not physically meaningful since MAE cannot be negative. Therefore, for clarity, the portion of the error bar below zero has been omitted. On the basis of average MAE, the sine-based grid shown in black has the least error, making it the best-performing grid for the simple harmonic oscillator in this setup. This also has a lower standard deviation of MAEs, indicating more consistent performance across various seeds. This can be further confirmation of the previous discussion, that the well clustering of points in the sine-based grid has made the PINN model easier to learn the oscillating nature of the solution by providing a sufficient number of training points in the crest and trough regions.

\begin{table}[!h] 
\centering
\caption{Training loss and evaluation MAE values for different random seeds and both Chebyshev/sine-based training point distributions to approximate the simple harmonic oscillator (Sec. \ref{Sec:RandomSeedSHO}). \label{Tab:SHOvalues}}
\begin{tabular}{c|c|c|c|c}
\textbf{Random} & \textbf{Chebyshev} & \textbf{Chebyshev} & \textbf{Sine based} &\textbf{Sine based}\\
\textbf{seed} & \textbf{loss} & \textbf{MAE} & \textbf{loss} &\textbf{MAE}\\
\hline
1 & $3.22\times 10^{-5}$ & $4.48\times 10^{-3}$ & $2.38\times 10^{-5}$ & $3.95\times 10^{-3}$ \\
11 & $1.54\times 10^{-6}$ & $9.87\times 10^{-4}$ & $1.34\times 10^{-6}$ & $9.28\times 10^{-4}$ \\		
21 & $2.01\times 10^{-5}$ & $3.58\times 10^{-3}$ & $3.38\times 10^{-6}$ & $1.54\times 10^{-3}$ \\		
31 & $1.36\times 10^{-5}$ & $2.87\times 10^{-3}$ & $4.74\times 10^{-6}$ & $1.85\times 10^{-3}$ \\	
41 & $1.53\times 10^{-5}$ & $2.85\times 10^{-3}$ & $1.09\times 10^{-5}$ & $2.65\times 10^{-3}$ \\	
51 & $8.40\times 10^{-6}$ & $2.29\times 10^{-3}$ & $1.56\times 10^{-6}$ & $1.04\times 10^{-3}$ \\	
61 & $6.93\times 10^{-4}$ & $2.20\times 10^{-2}$ & $4.62\times 10^{-5}$ & $5.27\times 10^{-3}$ \\	
71 & $5.11\times 10^{-5}$ & $5.54\times 10^{-3}$ & $1.29\times 10^{-6}$ & $9.32\times 10^{-4}$ \\
81 & $2.66\times 10^{-5}$ & $4.06\times 10^{-3}$ & $3.32\times 10^{-6}$ & $1.47\times 10^{-3}$ \\
91 & $7.95\times 10^{-6}$ & $2.22\times 10^{-3}$ & $1.48\times 10^{-6}$ & $1.03\times 10^{-3}$ \\

\end{tabular}
\end{table}

\subsubsection{Poisson equation}
\label{Sec:RandomSeedPoisson}

\begin{figure}[!h]
\centering
\includegraphics[width=1\linewidth]{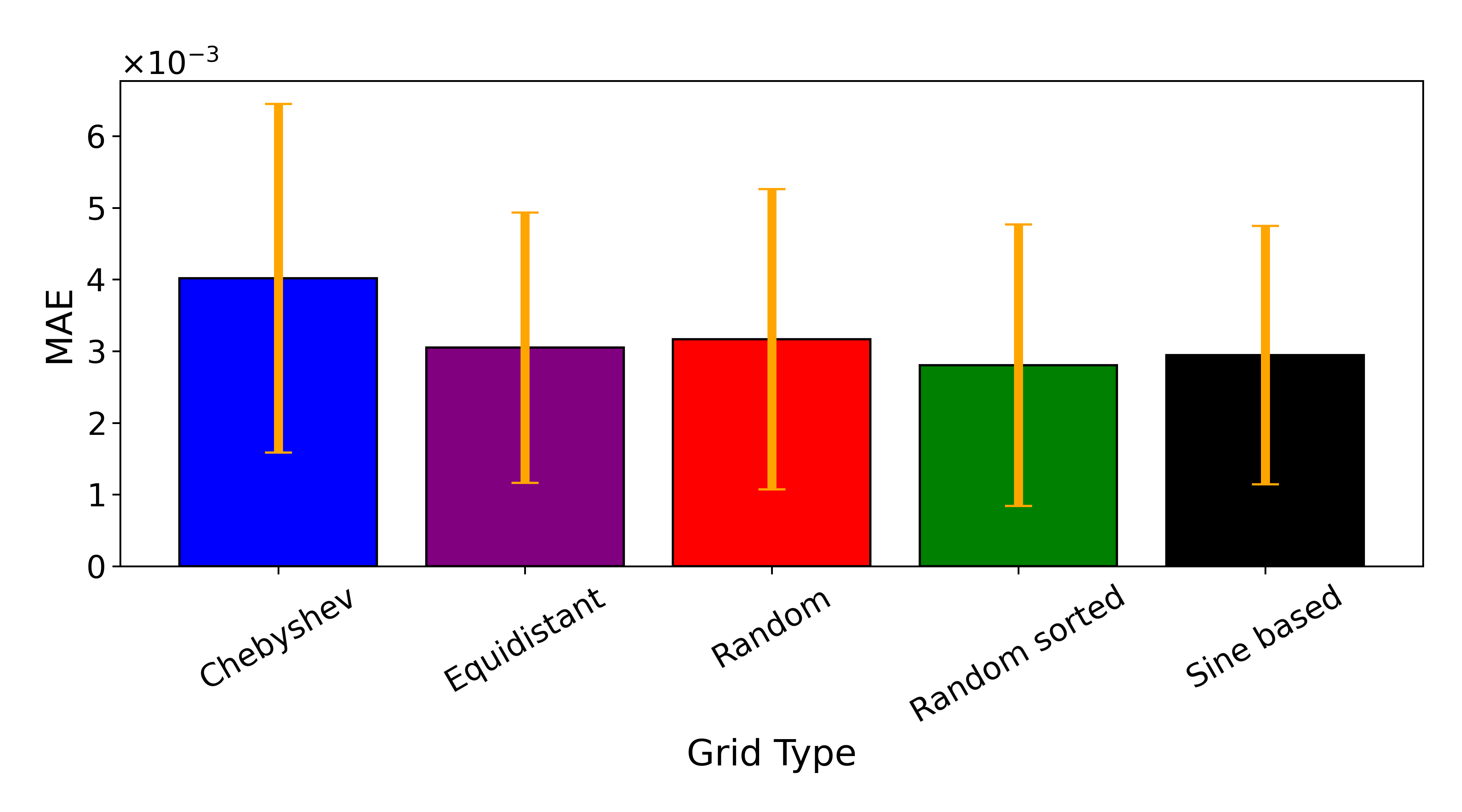}
\caption{Mean absolute errors averaged ($\overline{\text{MAE}}$ over 200 random seeds for the Poisson equation with a single layer FNN and 80$\times$80 training points (Sec.\ \ref{Sec:RandomSeedPoisson}). The bar colours refer to the $\overline{\text{MAE}}$ of each training point distribution strategy and the standard deviation (SD) is visualised in orange.}
\label{fig:Averaged_MAE_Poisson}
\end{figure}

Similar to the simple harmonic oscillator, the solution to the two-dimensional Poisson equation is predicted using the PINN model with one hidden layer and 100 neurons for 200 consecutive random seed initialisations. The model is trained using a grid size of $80\times80$ points. The objective is again to discuss the training point distribution strategies for an optimal setup, based on average MAE values from 200 random seeds.

Fig. \ref{fig:Averaged_MAE_Poisson} displays the average MAE values for different grid types, with standard deviations indicated by the orange lines on each bar. Among the different distributions, the random-sorted grid yields the lowest average MAE, suggesting that randomly selected training points can, on average, result in more accurate predictions. However, since random grids inherently vary between runs, there is always uncertainty in the outcome. Another randomly chosen set of training points, even under the same setup, may lead to different results.

\begin{table}[!h] 
\centering
\caption{Training loss and evaluation MAE values for different random seeds and both Chebyshev/sine-based training point distributions to approximate the Poisson equation (Sec. \ref{Sec:RandomSeedPoisson}). \label{Tab:Poisson_values}}
\begin{tabular}{c|c|c|c|c}
\textbf{Random} & \textbf{Chebyshev} & \textbf{Chebyshev} & \textbf{Sine based} &\textbf{Sine based}\\
\textbf{seed} & \textbf{loss} & \textbf{MAE} & \textbf{loss} &\textbf{MAE}\\
\hline
1 & $2.85\times 10^{-5}$ & $2.53\times 10^{-3}$ & $2.89\times 10^{-5}$ & $1.99\times 10^{-3}$ \\
11 & $2.18\times 10^{-5}$ & $2.12\times 10^{-3}$ & $3.73\times 10^{-5}$ & $4.29\times 10^{-3}$ \\		
21 & $3.18\times 10^{-5}$ & $3.25\times 10^{-3}$ & $5.15\times 10^{-5}$ & $2.72\times 10^{-3}$ \\		
31 & $6.06\times 10^{-5}$ & $5.66\times 10^{-3}$ & $3.95\times 10^{-5}$ & $2.49\times 10^{-3}$ \\	
41 & $3.36\times 10^{-5}$ & $3.79\times 10^{-3}$ & $1.87\times 10^{-5}$ & $1.15\times 10^{-3}$ \\	
51 & $4.12\times 10^{-5}$ & $4.39\times 10^{-3}$ & $7.03\times 10^{-5}$ & $4.77\times 10^{-3}$ \\	
61 & $3.40\times 10^{-5}$ & $3.66\times 10^{-3}$ & $3.79\times 10^{-5}$ & $2.33\times 10^{-3}$ \\	
71 & $2.99\times 10^{-5}$ & $2.15\times 10^{-3}$ & $1.52\times 10^{-5}$ & $1.02\times 10^{-3}$ \\
81 & $5.04\times 10^{-5}$ & $4.38\times 10^{-3}$ & $3.03\times 10^{-5}$ & $2.21\times 10^{-3}$ \\
91 & $1.08\times 10^{-4}$ & $1.89\times 10^{-3}$ & $3.32\times 10^{-5}$ & $2.47\times 10^{-3}$ \\

\end{tabular}
\end{table}

Tab. \ref{Tab:Poisson_values} presents the loss and MAE values at the final epoch for Chebyshev and sine-based distributions. This further clarifies that even when the training inputs and network architecture are the same, different initial weight values can lead to different outcomes. It highlights the significant role that weight initialisation plays in the optimisation process and the final performance of the FNN. 

Among the deterministic distribution, the equidistant grid yields the lowest error, with a standard deviation also lower than the others, indicating that this grid type is both more accurate and consistent. This is followed by the sine-based and Chebyshev grids. Interestingly, the Chebyshev grid resulted in the highest error for the Poisson PINN, which is contrary to our expectation. We had anticipated better accuracy from this grid, as the clustering of points near the boundaries was expected to help the network learn the boundary behaviour more effectively. 

\section{Discussion and future work}

In this paper we investigated and evaluated the performance of the PINN framework with five different strategies for training point distributions. With two ODEs and two PDEs for benchmark testing, we employed equidistant points, random and random sorted points, Chebychev points and a sine-base distribution for the training process. Each scenario was then evaluated using an equidistant distribution to capture the performance at grid points that are intermediate to all the training point setups. 

There is evidence that a careful choice of the training point distribution based on the characteristics and complexity of the differential equation can lead to better accuracy. For the radioactive decay, an initial value problem with rapid change at the start, the Chebyshev distribution tends to be the best strategy. It has a higher density at the boundaries and can therefore capture the initial state. The sine-based distribution should be favoured for differential equations with oscillatory behaviour, like the simple harmonic oscillator. This distribution aligns naturally with difficult parts in the oscillations. Regarding the tested partial differential equations, our results show a tendency towards the use of equidistant or random training point distribution. While in general larger FNNs are favoured in the literature, we cannot make a distinct statement on whether employing one or two hidden layers (with the chosen parameter setup) is favourable.

However, neural networks in general have many parameters that require fine-tuning for optimal results. This also holds for the PINN framework and although there is evidence of a correlation between the characteristics of a differential equation and the training point distribution strategy, we also observed that changing, e.g., the number of training points or the FNN architecture, almost always changed the results. 

Future work will include mixed distributions of, e.g., equidistant points with densely packed points at the boundaries and approaches towards more adaptive training point strategies. In addition, investigating larger FNN architectures and numerous differential equations may provide more information about preferable distribution strategy choices. 

\section*{Funding}
This work was funded by the Federal Ministry of Research, Technology and Space within the project \emph{KI@MINT} ("AI-Lab").

\section*{Data availability}
The authors declare, that upon reasonable request, the code is available from the corresponding author.

%
%
\printbibliography

@article{schneidereit2022computational,
  title={Computational characteristics of feedforward neural networks for solving a stiff differential equation},
  author={Schneidereit, T. and Breu{\ss}, M.},
  journal={Neural Computing and Applications},
  volume={34},
  number={10},
  pages={7975--7989},
  year={2022},
  publisher={Springer},
  doi={https://doi.org/10.1007/s00521-022-06901-6}
}

@phdthesis{schneidereit2022solution,
  author       = {Schneidereit, T.},
  title        = {The solution of time-dependent ordinary differential equations using neural networks: collocation polynomial neural forms and adaptive neural domain refinement},
  year         = {2022},
  school       = {BTU Cottbus-Senftenberg},
  type         = {Doctoral dissertation},
}

@inproceedings{han2020better,
  title={Better bootstrapping for approximate homomorphic encryption},
  author={Han, K. and Ki, D.},
  booktitle={Cryptographers’ Track at the RSA Conference},
  pages={364--390},
  year={2020},
  organization={Springer}
}

@book{boyce2017elementary,
  title     = {Elementary Differential Equations and Boundary Value Problems},
  author    = {Boyce, W. E. and DiPrima, R. C. and Meade, D. B.},
  edition   = {11th},
  isbn      = {9781119443766},
  lccn      = {2023302525},
  url       = {https://books.google.de/books?id=SyaVDwAAQBAJ},
  year      = {2017},
  publisher = {Wiley}
}

@book{young2012sears,
  title     = {Sears and Zemansky’s University Physics: With Modern Physics},
  author    = {Young, H. D. and Freedman, R. A. and Ford, A. L.},
  isbn      = {978-81-317-9027-4},
  edition   = {13th},
  year      = {2012},
  publisher = {Pearson Education}
}

@book{arfken2012mathematical,
  title     = {Mathematical Methods for Physicists: A Comprehensive Guide},
  author    = {Arfken, G. B. and Weber, H. J. and Harris, F. E.},
  isbn      = {978-0-12-384654-9},
  edition   = {7th},
  year      = {2012},
  publisher = {Academic Press}
}

@book{strauss2007partial,
  title     = {Partial Differential Equations: An Introduction},
  author    = {Strauss, W. A.},
  isbn      = {978-0470-05456-7},
  edition   = {2nd},
  year      = {2007},
  publisher = {John Wiley \& Sons}
}

@book{canale2014numerical, 
    title={Numerical methods for engineers}, 
    publisher={McGraw-Hill Education}, 
    author={Canale, R. P. and Chapra, S. C.},
    isbn  = {978-0-07-339792-4},
  edition = {7th},
    year={2014}}

@article{raissi2019physics,
  title     = {Physics-informed neural networks: A deep learning framework for solving forward and inverse problems involving nonlinear partial differential equations},
  author    = {Raissi, M. and Perdikaris, P. and Karniadakis, G. E.},
  journal   = {Journal of Computational Physics},
  volume    = {378},
  pages     = {686--707},
  year      = {2019},
  publisher = {Elsevier}
}

@article{lagaris1998artificial,
  title={Artificial neural networks for solving ordinary and partial differential equations},
  author={Lagaris, I. E. and Likas, A. and Fotiadis, D. I.},
  journal={IEEE transactions on neural networks},
  volume={9},
  number={5},
  pages={987--1000},
  year={1998},
  publisher={IEEE}
}

@article{lagaris2000neural,
  title={Neural-network methods for boundary value problems with irregular boundaries},
  author={Lagaris, I. E. and Likas, A. C. and Papageorgiou, D. G.},
  journal={IEEE Transactions on Neural Networks},
  volume={11},
  number={5},
  pages={1041--1049},
  year={2000},
  publisher={IEEE}
}

@article{cuomo2022scientific,
  title={Scientific machine learning through physics--informed neural networks: Where we are and what’s next},
  author={Cuomo, S. and Di Cola, V. S. and Giampaolo, F. and Rozza, G. and Raissi, M. and Piccialli, F.},
  journal={Journal of Scientific Computing},
  volume={92},
  number={3},
  pages={88},
  year={2022},
  publisher={Springer}
}

@article{baty2024hands,
  title={A hands-on introduction to Physics-Informed Neural Networks for solving partial differential equations with benchmark tests taken from astrophysics and plasma physics},
  author={Baty, H.},
  journal={arXiv preprint arXiv:2403.00599},
  year={2024}
}

@article{anadaptiveliu2024,
  title={An adaptive sampling method based on expected improvement function and residual gradient in PINNs},
  author={Liu, Y. and Chen, L. and Ding, J. and Chen, Y.},
  journal={IEEE Access},
  year={2024},
  publisher={IEEE}
}

@article{breuss2016numerical,
  title={A numerical study of Newton interpolation with extremely high degrees},
  author={Breu{\ss}, M. and Kemm, F. and Vogel, O.},
  journal={arXiv preprint arXiv:1609.08839},
  year={2016}
}

@article{fang2023physics,
  title={A physics-informed neural network based on mixed data sampling for solving modified diffusion equations},
  author={Fang, Q. and Mou, X. and Li, S.},
  journal={Scientific Reports},
  volume={13},
  number={1},
  pages={2491},
  year={2023},
  publisher={Nature Publishing Group UK London}
}

@article{guo2022novel,
  title={A novel adaptive causal sampling method for physics-informed neural networks},
  author={Guo, J. and Wang, H. and Hou, C.},
  journal={arXiv preprint arXiv:2210.12914},
  year={2022}
}

@article{rethinking_daw2022,
  title={Rethinking the importance of sampling in physics-informed neural networks},
  author={Daw, A. and Bu, J. and Wang, S. and Perdikaris, P. and Karpatne, A.},
  journal={arXiv preprint arXiv:2207.02338},
  year={2022}
}

@article{kaplarevic2023identifying,
  title={Identifying optimal architectures of physics-informed neural networks by evolutionary strategy},
  author={Kaplarevi{\'c}-Mali{\v{s}}i{\'c}, A. and Andrijevi{\'c}, B. and Bojovi{\'c}, F. and Nikoli{\'c}, S. and Krsti{\'c}, L. and Stojanovi{\'c}, B. and Ivanovi{\'c}, M.},
  journal={Applied Soft Computing},
  volume={146},
  pages={110646},
  year={2023},
  publisher={Elsevier}
}

@inproceedings{saratchandran2024activation,
  title={From activation to initialization: Scaling insights for optimizing neural fields},
  author={Saratchandran, H. and Ramasinghe, S. and Lucey, S.},
  booktitle={Proceedings of the IEEE/CVF Conference on Computer Vision and Pattern Recognition},
  pages={413--422},
  year={2024}
}

@inproceedings{batuwatta2023weight,
  title={Weight initialization in physics-informed neural networks to enhance consistency of mass-loss predictions of plant cells undergoing drying},
  author={Batuwatta-Gamage$^1$, C. and Rathnayaka, C. and Karunasena, H. C. P. and Karim, M. and Gu$^1$, Y.},
  booktitle={Proceedings of the International Conference on Computational Methods},
  volume={10},
  pages={186--196},
  year={2023}
}

@article{kingma2014adam,
  title={Adam: A method for stochastic optimization},
  author={Kingma, D. P. and Ba, J.},
  journal={arXiv preprint arXiv:1412.6980},
  year={2014}
}

@article{dubey2022activation,
  title={Activation functions in deep learning: A comprehensive survey and benchmark},
  author={Dubey, S. R. and Singh, S. K. and Chaudhuri, B. B.},
  journal={Neurocomputing},
  volume={503},
  pages={92--108},
  year={2022},
  publisher={Elsevier}
}

@article{Wang2023AnEG,
  title={An Expert's Guide to Training Physics-informed Neural Networks},
  author={Wang, S. and Sankaran, S. and Wang, H. and Perdikaris, P.},
  journal={ArXiv},
  year={2023},
  volume={abs/2308.08468},
  url={https://api.semanticscholar.org/CorpusID:260925531}
}

@article{Epperson01041987,
author = {Epperson, J. F.},
title = {On the Runge Example},
journal = {The American Mathematical Monthly},
volume = {94},
number = {4},
pages = {329--341},
year = {1987},
publisher = {Taylor \& Francis},
doi = {10.1080/00029890.1987.12000642}}

@article{JMLR:v18:17-468,
  author  = {Baydin, A. G. and Pearlmutter, B. A. and Radul, A. A. and Siskind, J. M.},
  title   = {Automatic Differentiation in Machine Learning: a Survey},
  journal = {Journal of Machine Learning Research},
  year    = {2018},
  volume  = {18},
  number  = {153},
  pages   = {1--43},
  url     = {http://jmlr.org/papers/v18/17-468.html}
}

@inproceedings{AutoDiff,
  title={Automatic differentiation in PyTorch},
  author={Paszke, A. and Gross, S. and Chintala, S. and Chanan, G. and Yang, E. and De-Vito, Z. and Lin, Z. and Desmaison, A. and Antiga, L. and Lerer, A.},
  booktitle={Proceedings of the 31st Conference on Neural Information Processing Systems},
  year={2017}
}

@article{algoritmy,
author = {Schneidereit, T. and Michael Breuß, M.},
title = {Solving Ordinary Differential Equations using Artificial Neural Networks - A study on the solution variance},
journal = {Proceedings of the Conference Algoritmy},
year = {2020},
pages = {21--30},	
url ={http://www.iam.fmph.uniba.sk/amuc/ojs/index.php/algoritmy/article/view/1547}
}

@misc{tensorflow2015-whitepaper,
title={{TensorFlow}: Large-Scale Machine Learning on Heterogeneous Systems},
url={https://www.tensorflow.org/},
note={Software available from tensorflow.org},
author={
    Mart\'{i}n~Abadi and
    Ashish~Agarwal and
    Paul~Barham and
    Eugene~Brevdo and
    Zhifeng~Chen and
    Craig~Citro and
    Greg~S.~Corrado and
    Andy~Davis and
    Jeffrey~Dean and
    Matthieu~Devin and
    Sanjay~Ghemawat and
    Ian~Goodfellow and
    Andrew~Harp and
    Geoffrey~Irving and
    Michael~Isard and
    Yangqing Jia and
    Rafal~Jozefowicz and
    Lukasz~Kaiser and
    Manjunath~Kudlur and
    Josh~Levenberg and
    Dandelion~Man\'{e} and
    Rajat~Monga and
    Sherry~Moore and
    Derek~Murray and
    Chris~Olah and
    Mike~Schuster and
    Jonathon~Shlens and
    Benoit~Steiner and
    Ilya~Sutskever and
    Kunal~Talwar and
    Paul~Tucker and
    Vincent~Vanhoucke and
    Vijay~Vasudevan and
    Fernanda~Vi\'{e}gas and
    Oriol~Vinyals and
    Pete~Warden and
    Martin~Wattenberg and
    Martin~Wicke and
    Yuan~Yu and
    Xiaoqiang~Zheng},
  year={2015},
}

@article{scikit-learn,
  title={Scikit-learn: Machine Learning in {P}ython},
  author={Pedregosa, F. and Varoquaux, G. and Gramfort, A. and Michel, V.
          and Thirion, B. and Grisel, O. and Blondel, M. and Prettenhofer, P.
          and Weiss, R. and Dubourg, V. and Vanderplas, J. and Passos, A. and
          Cournapeau, D. and Brucher, M. and Perrot, M. and Duchesnay, E.},
  journal={Journal of Machine Learning Research},
  volume={12},
  pages={2825--2830},
  year={2011}
}

@ARTICLE{2020SciPy-NMeth,
  author  = {Virtanen, Pauli and Gommers, Ralf and Oliphant, Travis E. and
            Haberland, Matt and Reddy, Tyler and Cournapeau, David and
            Burovski, Evgeni and Peterson, Pearu and Weckesser, Warren and
            Bright, Jonathan and {van der Walt}, St{\'e}fan J. and
            Brett, Matthew and Wilson, Joshua and Millman, K. Jarrod and
            Mayorov, Nikolay and Nelson, Andrew R. J. and Jones, Eric and
            Kern, Robert and Larson, Eric and Carey, C J and
            Polat, {\.I}lhan and Feng, Yu and Moore, Eric W. and
            {VanderPlas}, Jake and Laxalde, Denis and Perktold, Josef and
            Cimrman, Robert and Henriksen, Ian and Quintero, E. A. and
            Harris, Charles R. and Archibald, Anne M. and
            Ribeiro, Ant{\^o}nio H. and Pedregosa, Fabian and
            {van Mulbregt}, Paul and {SciPy 1.0 Contributors}},
  title   = {{{SciPy} 1.0: Fundamental Algorithms for Scientific
            Computing in Python}},
  journal = {Nature Methods},
  year    = {2020},
  volume  = {17},
  pages   = {261--272},
  adsurl  = {https://rdcu.be/b08Wh},
  doi     = {10.1038/s41592-019-0686-2},
}
\end{document}